\documentclass{article}
\usepackage{graphicx}
\usepackage{subcaption}
\usepackage{authblk}
\usepackage{amsmath}
\usepackage{multicol}
\usepackage{multirow}
\usepackage{caption}
\usepackage{url} 
\usepackage{makecell}
\usepackage{array}
\usepackage{bbm}
\usepackage{booktabs}
\usepackage{algorithm}
\usepackage{algpseudocode}
\usepackage{hyperref}
\usepackage{listings}
\usepackage{xcolor}

\lstdefinelanguage{YAML}{
  morekeywords={true,false,null,y,n},
  sensitive=false,
  morecomment=[l]{\#},
  morestring=[b]",
}

\usepackage[letterpaper, total={7.5in, 9.5in}]{geometry}
\usepackage[font=small,labelfont=bf]{caption}
\title{CEHR-XGPT: A Scalable Multi-Task Foundation Model for Electronic Health Records}

\author[1, 2]{Chao Pang} 
\author[3]{Jiheum Park}
\author[1, 2]{Xinzhuo Jiang}
\author[1, 2]{Nishanth Parameshwar Pavinkurve}
\author[1, 2]{Krishna S. Kalluri}
\author[1, 2]{Shalmali Joshi}
\author[1, 2, 4, 5]{Noémie Elhadad}
\author[1, 2, 4]{Karthik Natarajan}
\affil[1]{Department of Biomedical Informatics, Columbia University Irving Medical Center}
\affil[2]{Observational Health Data Sciences and Informatics}
\affil[3]{Department of Medicine, Columbia University Irving Medical Center}
\affil[4]{Medical Informatics Services, NewYork-Presbyterian Hospital}
\affil[5]{Department of Computer Science, Columbia University}

\date{}

\newcommand{\ehrshot}{\textit{ehrshot}}
\newcommand{\halo}{\textbf{HALO}}
\newcommand{\ethos}{\textbf{ETHOS}}
\newcommand{\foresight}{\textbf{Foresight}}
\newcommand{\cehrgpt}{\textbf{CEHR-XGPT}}
\newcommand{\gptvanilla}{\textbf{GPT}}
\newcommand{\motor}{\textbf{MOTOR}}
\newcommand{\clmber}{\textbf{CLMBR}}
\newcommand{\llama}{\textbf{Llama}}
\newcommand{\mamba}{\textbf{Mamba}}
\newcommand{\gpt}{\textbf{GPT2}}

\begin{document}

\maketitle

\begin{multicols}{2}
\raggedcolumns

\section*{Abstract}
Electronic Health Records (EHRs) provide a rich, longitudinal view of patient health and hold significant potential for advancing clinical decision support, risk prediction, and data-driven healthcare research. However, most artificial intelligence (AI) models for EHRs are designed for narrow, single-purpose tasks, limiting their generalizability and utility in real-world settings. Here, we present CEHR-XGPT, a general-purpose foundation model for EHR data that unifies three essential capabilities - feature representation, zero-shot prediction, and synthetic data generation - within a single architecture. To support temporal reasoning over clinical sequences, CEHR-XGPT incorporates a novel time-token-based learning framework that explicitly encodes patients' dynamic timelines into the model structure. CEHR-XGPT demonstrates strong performance across all three tasks and generalizes effectively to external datasets through vocabulary expansion and fine-tuning. Its versatility enables rapid model development, cohort discovery, and patient outcome forecasting without the need for task-specific retraining.

\section*{Keywords}
EHR Foundation Model, Synthetic Electronic Health Records, Patient Representation, Observational Medical Outcomes Partnership - Common Data Model, Observational Health Data Sciences and Informatics

\section{Introduction}
Electronic health records (EHRs) capture rich, longitudinal information about patients across diverse clinical settings. This wealth of data presents significant opportunities to build generalizable and clinically useful AI models. However, it is highly challenging to model such data reliably. EHRs differ from typical sequential data in several important ways: they are irregularly sampled, temporally sparse, and often contain heterogeneous modalities such as diagnoses, procedures, lab results, and medications. Accurately modeling the temporality and structure of EHR data is essential to capture the underlying progression of patient health and produce reliable, context-aware predictions \cite{Hripcsak2013, Hripcsak2015, Hripcsak2016_drug}. 

With the advent of general-purpose foundation models, such as ChatGPT—trained on large, diverse datasets using self-supervised learning rather than single-task objectives—researchers have begun developing analogous foundation models for EHRs \cite{Bommasani2021, Wornow2024}. These EHR foundation models aim to capture the full potential of longitudinal patient data by learning generalizable representations that can support a variety of downstream tasks with improved performance.

Recent EHR foundation models have been applied to specific applications such as disease prediction, patient outcome forecasting, synthetic data generation, extracting patient representations, and propensity score modeling for causal effect estimation \cite{Theodorou2023, Steinberg2020, Steinberg2023, Wornow2023, Wornow2024, Rasmy2021, Pang2021, Renc2024, Poulain2022, Fallahpour2024, McDermott2023, Rao2024}. However, most existing models have been designed and evaluated with a narrow focus—typically excelling at one task type, such as representation learning or predictive modeling, but not both.

We present \cehrgpt{} (pronounced “seer-ex-gpt”), a unified EHR foundation model that supports three core capabilities within a single framework: feature representation, zero-shot prediction, and synthetic data generation. By integrating all three, our model demonstrates a broader and more flexible utility compared to prior approaches.

\begin{itemize}
    \item \textbf{Feature Representation}: The model generates patient embeddings from sequences of medical events, enabling a variety of downstream tasks such as disease prediction, patient clustering, and propensity score matching.
    
    \item \textbf{Zero-Shot Prediction}: The model can predict future patient events directly from prompts, without requiring task-specific training or fine-tuning. This enables rapid evaluation of new prediction tasks in settings with limited labeled data.
    
    \item \textbf{Synthetic Data Generation}: The model learns the sequence distribution of real patient data and generates synthetic timelines that preserve key statistical properties and inter-variable dependencies, supporting tasks like simulation, data sharing, and augmentation.
\end{itemize}

The versatility of \cehrgpt{} stems from a novel time-token–based learning framework that explicitly encodes patients’ dynamic timelines into the model architecture, preserving the full temporal structure and sequence dependencies inherent in longitudinal EHR data. To our knowledge, \cehrgpt{} is the first multi-task foundation model to simultaneously support all three core capabilities within a unified framework.






\section{Related work}
There has been a growing number of studies on EHR foundation models, each targeting slightly different capabilities, including feature representation, zero-shot prediction, and synthetic data generation. 

\textbf{Feature Representation}: Recent works such as \motor{}, along with models like \llama{} and \mamba{} reported by Wornow \textit{et al.}, have focused on the feature representation capabilities of EHR foundation models using linear probing \cite{Steinberg2020, Steinberg2023, Wornow2024}. Linear probing is a model evaluation technique in which a lightweight classifier (typically logistic regression) is trained on fixed patient representations generated by a pretrained foundation model. This approach allows researchers to assess how well the model's learned embeddings encode clinically relevant information, without further training for specific tasks. In the context of EHR applications, linear probing offers several advantages: it requires only a single forward pass through the model, is computationally efficient, and avoids the need for fine-tuning, making it attractive for rapid evaluation and deployment. In contrast, other models such as \textbf{MED-BERT}, \textbf{BEHRT}, \textbf{TransformerEHR}, and \textbf{EHRMamba} require additional fine-tuning for each downstream task after pretraining \cite{Pang2021, Yang2023, Rasmy2021, Li2020, Poulain2022, Fallahpour2024, Li2023, Odgaard2024, Hur2023}. These models differ in two main aspects. First, they adopt various architectural choices, including encoder-only, decoder-only, and encoder–decoder designs. Second, they incorporate temporal information in different ways, such as through positional embeddings, age embeddings, time embeddings, or, more recently, discrete time tokens. 

\textbf{Zero-Shot Prediction}: Several EHR Foundation Models, such as \foresight{} and \ethos{}, built on causal language model architectures, focus on zero-shot prediction without any task-specific training \cite{Kraljevic2022, McDermott2023, Renc2024}. In Foresight, the authors first used NLP tools to extract clinical concept codes from unstructured notes, which they combined with structured EHR data to construct full patient timelines. To encode demographic context, age tokens were inserted into the timeline, with a new token added whenever the patient’s age increased. \foresight{} achieved strong performance in forecasting future conditions, demonstrating high precision and recall across ranked predictions (e.g., top-1, top-5, top-10). A qualitative evaluation was also conducted, in which five clinicians assessed predictions generated from 34 synthetically constructed patient timelines. \foresight{} achieved a high relevance score, with 33 out of 34 predictions (95\%) deemed clinically meaningful. Similarly, \ethos{}, trained primarily on MIMIC-IV, demonstrated strong zero-shot capabilities, matching or surpassing prior SOTA models on several clinical prediction benchmarks. However, both models employed tokenization strategies that compromised temporal fidelity. In \foresight{}, age tokens served as coarse temporal markers, making it impossible to distinguish between events occurring 1 day apart versus 90 days apart if the patient's age remained constant. In \ethos{}, more granular time tokens (e.g., 6–12 hours, 3–6 months) were introduced to better model ICU timelines, but this approach still struggled to capture time gaps across discrete hospital admissions. While such models are well-suited for forecasting tasks that tolerate temporal compression, they are less appropriate for applications like synthetic data generation or fine-grained trajectory modeling, which require high temporal resolution.

\textbf{Synthetic Data Generation}: Structured EHRs are characterized as irregular time-series data with high-dimensional features at each time point. Most prior work has focused on deep learning-based methods without adequately preserving the temporal structure of clinical events \cite{Choi2017, Li2017, Cui2019, Baowaly2019, Yan2020, Lee2020, Torfi2020, Rashidian2020, Biswal2021, Sun2021, Zhang2021, Theodorou2023, Li2023, Yoon2023}. For example, \halo{}, a state-of-the-art (SOTA) model for generating synthetic EHR data, incorporates inter-visit time intervals but fails to represent inpatient visit durations \cite{Theodorou2023}. This omission makes it unsuitable for tasks that require accurate modeling of patient trajectories over time, such as identifying 30-day readmissions, thereby limiting its evaluation to simpler tasks like predicting the presence of medical codes rather than supporting more complex phenotype modeling. Moreover, \halo{} discretizes inter-visit intervals into coarse buckets and samples from a uniform distribution, which can distort realistic patient timelines. In real-world outpatient care, for instance, 7- and 14-day intervals are much more common. Finally, \halo{} was evaluated separately on inpatient and outpatient datasets rather than on a unified EHR dataset that spans both care settings, further limiting its generalizability.

These capabilities require varying levels of patient timeline preservation and generative capacity. For \textbf{feature representation}, full patient timelines are not strictly necessary—some foundation models are not generative and simply order medical events chronologically without modeling temporal dependencies \cite{Li2020, Rasmy2021}. In \textbf{zero-shot prediction}, models are not required to reconstruct complete timelines; some degree of timeline shrinkage or abstraction can be tolerated without significantly impacting performance \cite{Kraljevic2022, McDermott2023}. In contrast, \textbf{synthetic data generation} requires high-fidelity modeling of entire patient trajectories. To generate realistic synthetic sequences, it is essential to preserve the full temporal structure and the sequence dependencies inherent in the original data \cite{Theodorou2023, Pang2024}. As a result, a strong synthetic data generator should, in theory, also perform well on zero-shot prediction and feature representation tasks; however, the reverse does not necessarily hold.

\begin{figure*}[!htbp]
    \vspace{-1em} 
    \centering
    \includegraphics[width=0.94\linewidth]{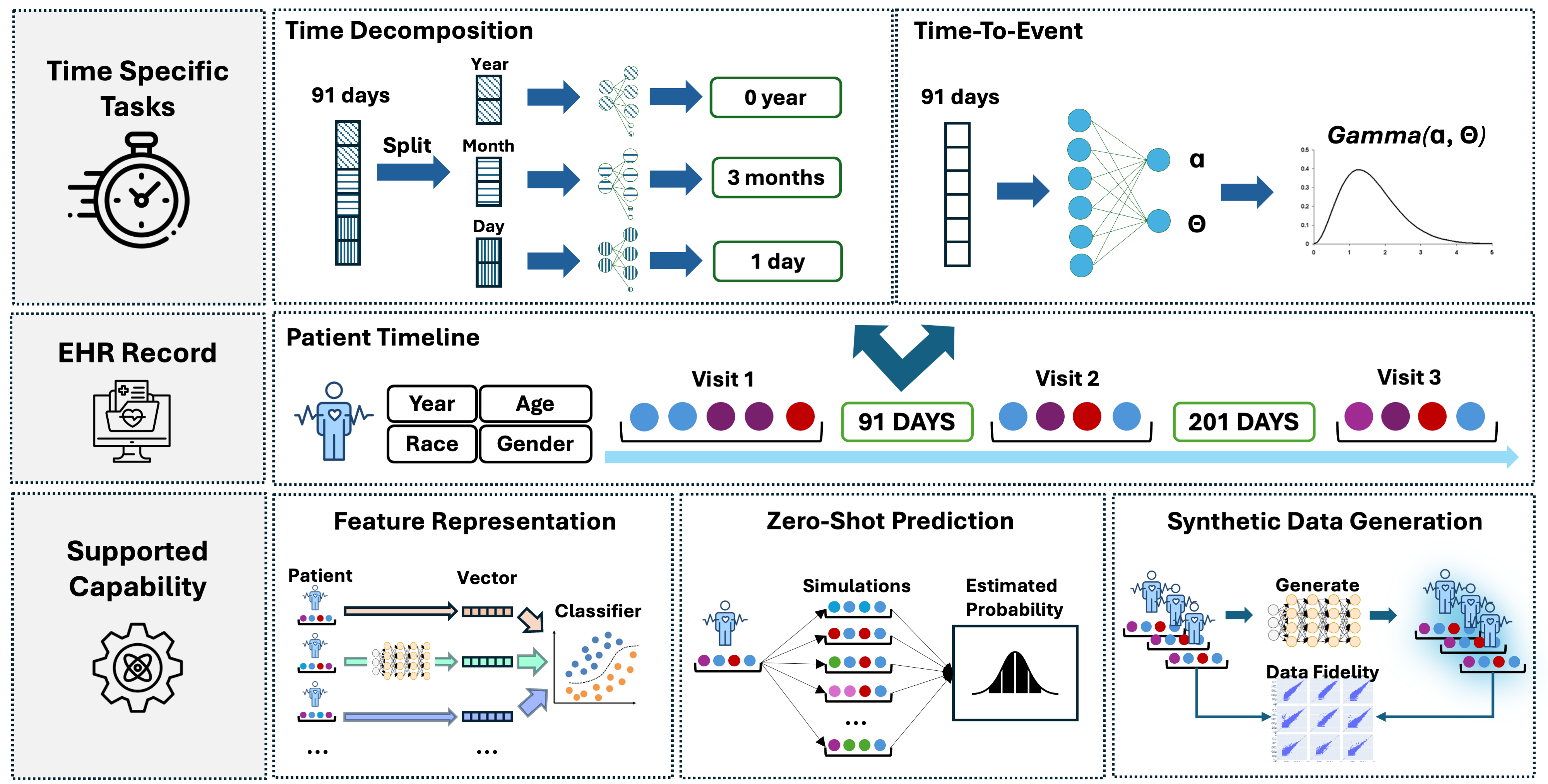}
    \caption{\cehrgpt{} is built on the GPT-2 architecture with a specialized patient representation designed to capture the full longitudinal trajectory of each patient. Two additional learning objectives were applied to artificial time token embeddings to improve the modeling of temporal information, namely, time decomposition and time-to-event prediction. \cehrgpt{} supports three key capabilities: patient-level feature representation, zero-shot prediction, and synthetic data generation.}
    \label{overview}
    \vspace{-1em} 
\end{figure*}

\section{Methods}

\subsection{Data Source and Preprocessing} \label{Data Source}
The training patient sequences were derived from a subset of the Columbia University Irving Medical Center-New York Presbyterian Hospital OMOP database. This subset includes data on conditions, medications, and procedures from patient medical histories. Unknown concepts (i.e., \(\texttt{concept\_id} = 0\)) were excluded from all domains except the visit type when constructing the patient sequences using our proposed representation. Additionally, we excluded patients with fewer than 20 tokens to ensure sufficient longitudinal information for model training.  The final dataset included approximately 2.6 million patients for training and an additional 1 million patients reserved for evaluation. To enable comparison with other baselines such as \motor{} \cite{arnrich2024medical, Steinberg2023}, we also converted the CUIMC-NYP OMOP dataset into the Medical Event Data Standard (MEDS) format. 

\subsection{Representing Patient Trajectory} \label{Patient Sequence}
Each patient's EHR record is modeled as a sequence of events, where the demographic information (comprising \texttt{start\_year}, \texttt{start\_age}, \texttt{gender}, and \texttt{race}) is positioned at the beginning. This is followed by the patient's visits, arranged chronologically. Due to the irregular nature of patient trajectories, where time intervals between visits vary, we introduce artificial time tokens (ATT) to quantify these intervals in days. For intervals shorter than 1080 days, a corresponding ATT token, such as $D10$ for a 10-day gap, is used. Intervals exceeding 1080 days are denoted by a single \texttt{[LT]} token (long-term). Each visit is encapsulated between two special tokens, \texttt{[VS]} and \texttt{[VE]}, to mark the start and end of a visit, respectively. Immediately following \texttt{[VS]}, a visit-type token \texttt{[VT]} specifies the type of visit. For inpatient visits, a discharge-type token is inserted at the end of the visit right before \texttt{[VE]}. The structure of a patient sequence is thus represented as:
\begin{align*}
    P = \{ &\texttt{[start\_year]}, \texttt{[start\_age]}, \texttt{[gender]}, \texttt{[race]} \\
    &\texttt{[VS] [VT]}, v_1, \texttt{[VE]}, ATT \\
    &\texttt{[VS] [VT]}, v_2, \texttt{[discharge\_type][VE]}, ATT \\
    &\dots \\
    &\texttt{[VS] [VT]}, v_i, \texttt{[VE]} \}
\end{align*}
where $v_i$ represents an arbitrary visit comprising a chronologically ordered list of medical events (denoted as $c_i$), and $v_i = \{c_{i1}, c_{i2}, \dots, c_{ij}\}$.

\subsection{Model Architecture} \label{Model Architecture}
We developed our model architecture based on the GPT-2 framework \cite{Radford}, tailored to effectively capture the sequence distribution. To enhance the model's efficiency, we excluded the positional embeddings from the model due to two primary considerations:
\begin{enumerate}
    \item The temporal information is inherently captured through the use of ATT tokens, making separate positional embeddings unnecessary.
    \item As demonstrated by Haviv et al~\cite{Haviv2022}, transformer-based language models such as GPT can learn and leverage positional information implicitly, even in the absence of explicit or structured positional encodings. The causal attention mechanism employed by GPT inherently enables the model to account for temporally ordered patient histories by conditioning on all prior tokens in a sequence, without requiring additional structured inputs such as positional embeddings.
\end{enumerate}
Our preliminary experiments supported this design choice, showing that trainable positional embeddings across different positions tended to converge to similar vectors, suggesting their limited utility in our setup. Additionally, we incorporated two specialized learning objectives, Time Decomposition (TD) and Time to Event (TTE), designed to enhance the representation of time intervals within the model. The overall architecture of the \cehrgpt{} foundation model is illustrated in Figure \ref{overview}.

\begin{figure}[H]
    \centering
    \includegraphics[width=\linewidth]{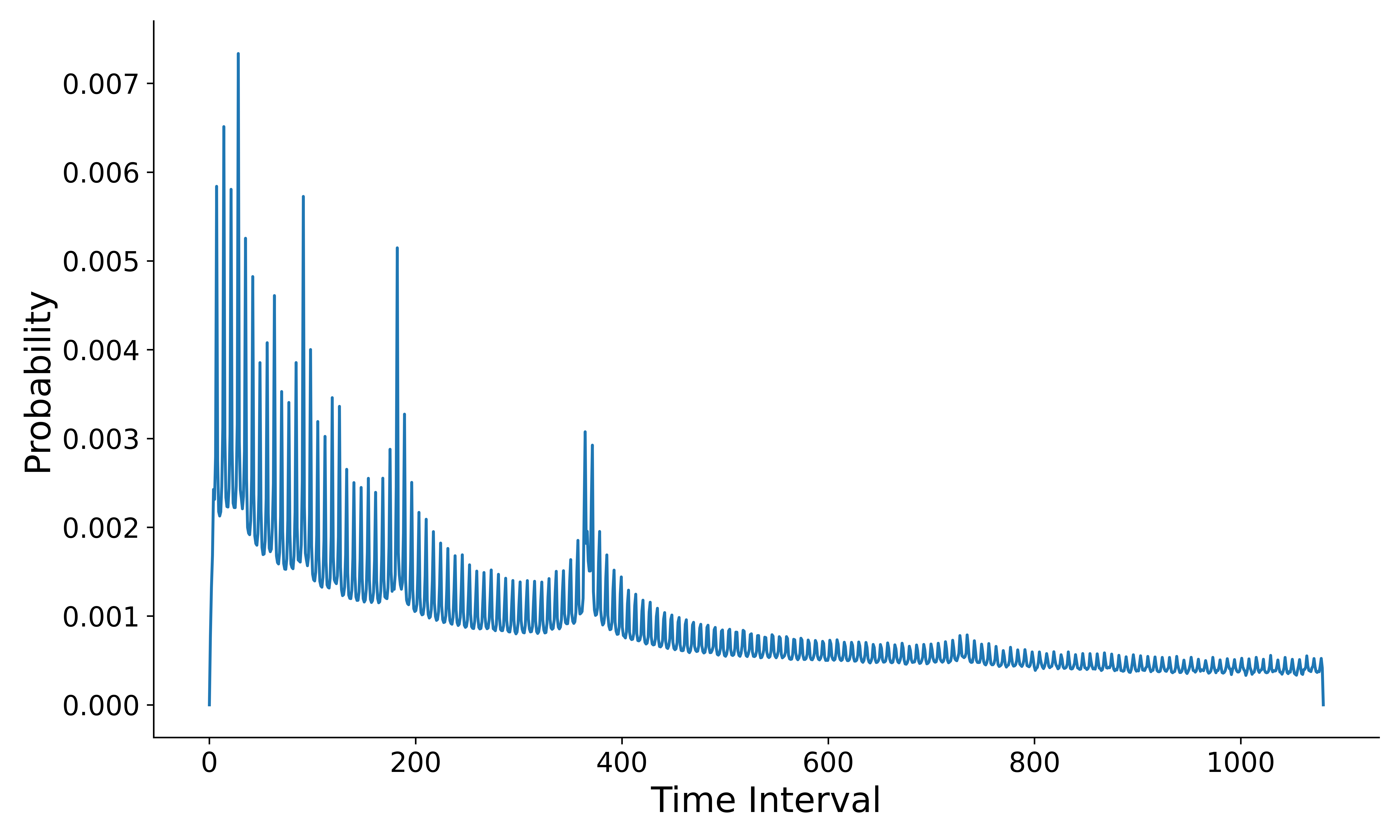}
    \caption{The distribution of time tokens between visits is not uniform but instead biased toward specific intervals, such as 7 and 14 days. }
    \label{time_token_distribution}
\end{figure}

The TD learning objective introduces a structural constraint on the ATTs to address the skewness in the distribution of time intervals, as depicted in Figure \ref{time_token_distribution}. While short-term intervals are more frequent, there is a noticeable long tail extending towards the long-term ATTs. The TD principle allows for the decomposition of any time interval into years, months, and days. For example, $D1$ would be decomposed into $0$ year, $0$ month, and $1$ day, whereas $D396$ would be translated to $1$ year, $1$ month, and $1$ day. 

To implement this, we added a TD prediction head at the positions corresponding to ATT tokens within the output layer. Each ATT embedding is split into three sub-embeddings that predict the time components of year, month, and day. This multiclass classification scenario uses cross-entropy loss to compute the loss term, allowing the model to leverage shared TD predictions across different timescales to improve the accuracy of long-term ATT predictions. The TD loss term can be expressed as follows,

\begin{align*}
    [e^{year}_{ij}&, e^{month}_{ij}, e^{day}_{ij}] = split(e^{att}_{ij}) \\
     &\hat{y}^{year}_{ij} = W_{year} \times e^{year}_{ij} \\
     &\hat{y}^{month}_{ij} = W_{month} \times e^{month}_{ij} \\
     &\hat{y}^{day}_{ij} = W_{day} \times  e^{day}_{ij} \\
    TD_{ij} =& - \sum P(y^{year}_{ij}) \log P(\hat{y}^{year}_{ij}  \mid e_{ij}^{year}) \\ 
   &- \sum P(y^{month}_{ij}) \log P(\hat{y}^{month}_{ij}  \mid e_{ij}^{month}) \\
   &- \sum P(y^{day}_{ij}) \log P(\hat{y}^{day}_{ij} \mid e_{ij}^{day})
\end{align*}

where $e^{att}_{ij}$ denotes the embedding representation of an ATT at the $jth$ position of the $ith$ patient, $e^{year}_{ij}$, $e^{month}_{ij}$, $e^{day}_{ij}$ represent the sub-embeddings corresponding to year, month and day. $W_{year}$, $W_{month}$, $W_{day}$ represent the linear maps to transform the sub-embeddings to the corresponding time component predictions.

In addition to incorporating TD, we introduced a Time-To-Event (TTE) learning objective to enhance the semantic depth of ATTs. When ATT embeddings are trained solely through next-token prediction, driven primarily by co-occurrence statistics, the model is limited to learning only the relative positions of ATTs in the embedding space, reducing their semantic richness. To address this limiation, TTE uses a Gamma probability distribution, parameterized by a feed-forward layer, at positions corresponding to ATTs. The model is optimized by minimizing the negative log-likelihood of the actual time intervals under a Gamma distribution. The resulting TTE loss term is defined as:
\begin{align*}
    \alpha_{ij}, \beta_{ij} = feed\_forward(e^{att}_{ij}) \\
    TTE_{ij} = -\log P(\Delta T_{ij}; Gamma(\alpha_{ij}, \beta_{ij}))
\end{align*}
where $e^{att}_{ij}$ denotes the embedding representation of an ATT at the $jth$ position of the $ith$ patient, $\alpha_{ij}$ and $\beta_{ij}$ are the parameters for the Gamma distribution generated by a feed-forward network, $\Delta T_{ij}$ is the time difference in days represented by $e^{att}_{ij}$.

The overall loss function of \cehrgpt{} is defined below, which comprises the original next token prediction (NTP), denoted as $L^{ntp}_{ij}$, and the two additional learning objectives defined above, 
\begin{align*}
    L = &\sum_{i}\sum_{j} \Big (L^{ntp}_{ij} + \mathbbm{1}[c_{ij} \in ATTs] \big(TTE_{ij} + TD_{ij}\big) \Big)
\end{align*}

\begin{center}
\captionsetup{justification=raggedright,singlelinecheck=false}  
  \begin{tabular}{c|cc}
  \toprule 
  & 
  \thead{No. of visits} & 
  \thead{No. of concepts}\\
  \midrule
  mean & 16 & 117\\
  std & 19 & 278\\
  min & 2 & 8\\
  25\% & 1 & 14\\
  50\% & 3 & 32\\
  75\% & 11 & 98\\
  99\% & 216 & 1775 \\
  max & 1560 & 9994\\
    \hline
  \end{tabular}
  \captionof{table}{Summary statistics of the CUIMC-NYP OMOP training data}
  \label{tab:summary-stats}
\end{center}

\subsection{Training Setup}
 The model uses a 4096-token context window, 16 transformer decoder blocks, 12 attention heads, and 768 dimensional embeddings and hidden units. A dropout rate of $0.1$ was applied to reduce overfitting. Input sequences exceeding 4096 tokens were truncated to fit the context window. Summary statistics of the training data are provided in Table \ref{tab:summary-stats}, and a full model configuration is listed in Supplementary Table \ref{tab:model_summary}. The model was trained for 10 epochs on a single NVIDIA A6000 GPU using sample packing, with a batch size of up to 16,384 tokens and an initial learning rate of 0.001. To enable recovery and intermediate evaluation without retraining from scratch, checkpoints were saved every 20,000 training steps (approximately per epoch).

\subsection{Time tokens V.S. Time embeddings} \label{time_comparison}
In this section, we provide a theoretical justification for inserting time tokens into patient sequences, which may yield better results than the traditional summation strategy of adding time embeddings to concept embeddings - a common approach in prior work \cite{Li2020, Yang2023, Fallahpour2024}. 

To illustrate the limitations of the summation approach, we consider a simple logical function where the output depends on the time interval between two binary events \(x_1, x_2 \in \{0, 1\}\), with timestamps \(t_1, t_2\), where \(t_1 \leq t_2\):

\begin{enumerate}
    \item If \(t_2 - t_1 \leq 7\), then \(y = \text{XOR}(x_1, x_2)\).
    \item If \(t_2 - t_1 > 7\), then \(y = \text{AND}(x_1, x_2)\).
\end{enumerate}

When using a token-based input like \(\vec{x}_{\text{token}}^T = [x_1, \Delta t, x_2]\), a simple feedforward neural network can be constructed to correctly route the computation through the appropriate logic gate based on the time interval \(\Delta t\). We demonstrate this using an example input of \(\vec{x}_{\text{token}}^T = [0, 6, 1]\), along with hand-designed weights and thresholds, showing that the model can successfully activate XOR or AND logic depending on the interval.

We define matrices, $W$, and bias vectors, $\beta^T$:
\[
W_1 = \begin{bmatrix}
1 & 0 & 0 & 1 & 0 & 0 \\
0 & -1 & 0 & 0 & 1 & 0 \\
0 & 0 & 1 & 0 & 0 & 1
\end{bmatrix}, \quad
\beta_1^T = [0, 7.5, 0, 0, -7.5, 0]
\]

The modified ReLU function is defined as:
\[
\text{ReLU}(x) = \begin{cases} 
0 & \text{if } x \leq 0 \\
1 & \text{if } x > 0
\end{cases}
\]

With the input \(\vec{x}_{\text{token}}^T = [0, 6, 1]\), we calculate:
\[
a^1 = \text{ReLU}(\vec{x}_{\text{token}} \cdot W_1 + \beta_1) = [0, 1, 1, 0, 0, 1]
\]
The vector \(a^1\) processes the inputs for XOR and AND operations, where the middle elements act as switches to activate the respective operations. We require another layer defined by \(W_2\) and \(\beta_2\):
\[
W_2 = \begin{bmatrix}
1 & 0 & 0 & 0 \\
1 & 1 & 0 & 0 \\
0 & 1 & 0 & 0 \\
0 & 0 & 1 & 0 \\
0 & 0 & 1 & 1 \\
0 & 0 & 0 & 1
\end{bmatrix}, \quad
\beta_2^T = [-1.5, -1.5, -1.5, -1.5]
\]
\[
a^2 = \text{ReLU}(a^1 \cdot W_2 + \beta_2) = [0, 1, 0, 0]^T
\]
The outputs of \(a^2\) direct the inputs to the XOR and AND gates, assuming that $XOR$, $AND$, and $OR$ are existing neural networks. The weights \(W_1, \beta_1, W_2, \beta_2\) used in the above example should work with all the other inputs, solving this time-dependent logical operator function. 
\[
y = \text{OR}\Big(\text{XOR}(a_2 \cdot \begin{bmatrix} 1 & 0 \\ 0 &  1 \\ 0 & 0 \\ 0 & 0 \end{bmatrix}), \text{AND}(a_2 \cdot \begin{bmatrix} 0 & 0 \\ 0 &  0 \\ 1 & 0 \\ 0 & 1 \end{bmatrix})\Big) = 1
\]

In contrast, the summation strategy adds time and concept embeddings before inputting them into the model, resulting in inputs like \(\vec{x}_{\text{sum}} = [x_1 + t_1, x_2 + t_2]\). This blending can collapse different input combinations into the same value. For example, the group \(x_1=0\), \(x_2=1\), \(t_1=0\), \(t_2=6\) would result in the same value as another group \(x_1=0\), \(x_2=0\), \(t_1=0\), \(t_2=7\), even though these inputs should produce different outputs. Once time and concept embeddings are summed, it becomes impossible to recover the original values, making it difficult for neural networks to learn time-sensitive logic. This limitation arises from the rigid functional form imposed by the summation strategy, which constrains the model's flexibility in representing and reasoning over time intervals While such exact collisions are rare in high-dimensional space, this example highlights a structural limitation of the summation approach that can impair learning in temporally complex tasks. 

We further support this argument with a simulation study, presented in Supplementary Section \ref{simulation_study}, which demonstrates that time tokens more effectively preserve time-sensitive logic across varying event intervals compared to the summation strategy.

\subsection{Downstream Prediction Tasks} \label{Cohort Definition}

We implemented the cohorts listed in Table \ref{tab:cohort} to support evaluations of synthetic data generation and feature representation. However, two cohorts—\textit{Discharge home death} and \textit{T2DM HF}—were excluded from the synthetic data evaluation because they require information from the death and measurement domains, which are not included in the current version of our synthetic dataset.

\begin{table*}[hbt]
  \centering
  \begin{tabular}{cp{5.5cm}cccc}
  \toprule
   \thead{Cohort} & \thead{Description} & \thead{Size} & \thead{Median Age} & \thead{Female (\%)} & \thead{Outcome (\%)} \\
  \midrule
  HF readmission & 30-day all-cause readmission in heart failure (HF) patients. & 138,337 & 71 & 49.80 & 25.7 \\
  
  Discharge home death & Mortality within 1 year since discharge to home. & 227,256 & 50 & 66.89 & 4.41 \\
  
  T2DM HF & Lifetime heart failure (HF) prediction since the initial diagnosis of type 2 diabetes mellitus (T2DM). & 130,169 & 61 & 50.72 & 9.96 \\
  
  Hospitalization & 2-year risk of hospitalization starting from the 3rd year since the initial entry into the EHR system. & 722,295 & 45 & 61.99 & 5.91 \\

  AFib ischemic stroke & 1-year risk of ischemic stroke within one year of initial diagnosis of Atrial fibrillation (AFib). & 48,314 & 71 & 48.36 & 3.50\\

  CAD CABG & Patients with an initial diagnosis of Coronary Artery Disease (CAD) received Coronary Artery Bypass Grafting (CABG) surgery within a year, excluding prior stent grafts. & 78,990 & 69 & 45.6 & 4.27\\
  \bottomrule
  \end{tabular}
  \caption{Definitions and cohort characteristics of prediction tasks}
  \label{tab:cohort}
\end{table*}

\subsection{Evaluation Setup} \label{Evaluation Setup}
We evaluated \cehrgpt{} for three capabilities: feature representation, zero-shot prediction, and synthetic data generation. To benchmark our model against existing approaches, we compare it with \motor{}, a SOTA EHR foundation model known for its strong performance in representation learning tasks \cite{Steinberg2023}. However, since \motor{} does not support zero-shot prediction or synthetic data generation, our comparison is limited to representation performance. Furthermore, we introduce an ablation baseline—\gptvanilla{}—which shares the same architecture as \cehrgpt{} but excludes the proposed TTE and TD learning objectives. This allows us to assess the contribution of our novel timeline design.

\subsubsection{Feature Representation}
We evaluated the feature representation capabilities of the models using a linear probing strategy, as proposed by Steinberg \textit{et al.} and  Wornow \textit{et al.} \cite{Steinberg2023, Wornow2024}. This approach involved extracting patient-level feature vectors from the final layer of the transformer decoder, which were then used to train a logistic regression classifier. These experiments were conducted on the disease prediction tasks described in Table \ref{tab:cohort}. For comparison, we included \gptvanilla{} and \motor{}, both pre-trained on the Columbia dataset. The same linear probing methodology was applied to these models. Additionally, we included a logistic regression baseline trained on bag-of-words (BOW) features, as described in Section \ref{synthetic_data_generation}.

Beyond linear probing, we also fine-tuned \cehrgpt{} and \gptvanilla{} for each prediction task. The training data was split into training and validation subsets using a 90:10 ratio, with the validation set used for early stopping to prevent overfitting. We performed hyperparameter tuning using 10\% of the training and validation data, running 10 trials with hyperparameters sampled uniformly: learning rates ranged from $1 \times 10^{-5}$ to $5 \times 10^{-5}$, and weight decay values ranged from 0.01 to 0.05. The optimal hyperparameters were selected based on the lowest validation loss. The models were then retrained on the full training set, using early stopping with a patience of 1 to monitor validation loss.  \motor{} was not fine-tuned, as Steinberg \textit{et al.} \cite{Steinberg2023} reported that its fine-tuned and linearly probed versions exhibit comparable performance. Final performance was evaluated on the held-out test set, and we report both AUROC and AUPRC metrics. The 95\% confidence intervals were computed using bootstrapping.

\subsubsection{Zero-Shot Prediction}
Our evaluation followed the zero-shot setup proposed by \ethos{} \cite{Renc2024}, in which the patient sequence up to the prediction time was provided to \cehrgpt{} to generate 50 simulated patient trajectories. Generation was terminated when either of the following conditions was met: (1) the generated tokens exceeded the defined prediction window, or (2) the outcome event(s) occurred within the prediction time. If the \textit{[END]} token was encountered within the prediction window—causing premature termination—we treated the simulation as censored and discarded it, replacing it with a new one. The probability of the predicted outcome was then estimated as the proportion of simulations in which the event occurred. Using these estimated probabilities and the corresponding ground-truth labels, we computed AUROC and AUPRC scores. The standard deviation of these estimates was assessed via bootstrapping. An illustrative example of this zero-shot prediction setup is provided in Supplementary Section~\ref{sec:zero_shot_setup}. Due to the computational cost associated with the generative nature of zero-shot inference, we randomly sampled 5,000 test cases from each prediction task. 

\subsubsection{Synthetic Data Generation}
We extended the synthetic data generation procedure developed by Pang \textit{et al.} \cite{Pang2024}, addressing one of the primary challenges identified in their work: optimizing the sampling strategy. To this end, we adopted a \textit{mixture of experts} approach, consisting of the following steps:

\begin{enumerate}
    \item \textbf{Data Generation:} Multiple synthetic datasets were generated using different combinations of sample sizes, model checkpoints, generation temperatures, \texttt{top\_p}, \texttt{top\_k}, and \texttt{repetition\_penalty} values.
    \item \textbf{Data Pooling and Conversion:} All synthetic patient sequences were pooled and converted into a unified OMOP instance.
\end{enumerate}

The resulting synthetic OMOP dataset contained approximately 3.7 million patient records, and details of this synthetic dataset can be found in Supplementary Section~\ref{sec:synthetic_data_evaluation}. Based on this dataset, we constructed four prediction tasks using methodologies described in \cite{Pang2021, Pang2024, ohdsi2019book}. Cohort definitions are summarized in Table~\ref{tab:cohort}.\medskip

We evaluated the utility of the synthetic data under two real-world simulation scenarios:

\begin{enumerate}
    \item \textbf{Model Development:} Training models solely on synthetic data.
    \item \textbf{Training Set Augmentation:} Augmenting real training data with synthetic data.
In both cases, performance was measured on a real-world test set.
\end{enumerate}

Feature extraction followed a bag-of-words (BOW) approach, counting concept frequency within defined observation windows. Logistic regression models were trained using Scikit-Learn’s default settings. For real data baselines, we used the original train/test splits.

In both cases, performance was measured on a real-world test set. Feature extraction followed a bag-of-words (BOW) approach, counting concept frequency within defined observation windows. Logistic regression models were trained using Scikit-Learn’s default settings. For real data baselines, we used the original train/test splits. The standard deviation was estimated through bootstrapping. 

\begin{figure*}[!htbp]
    \centering
    \includegraphics[width=0.95\linewidth]{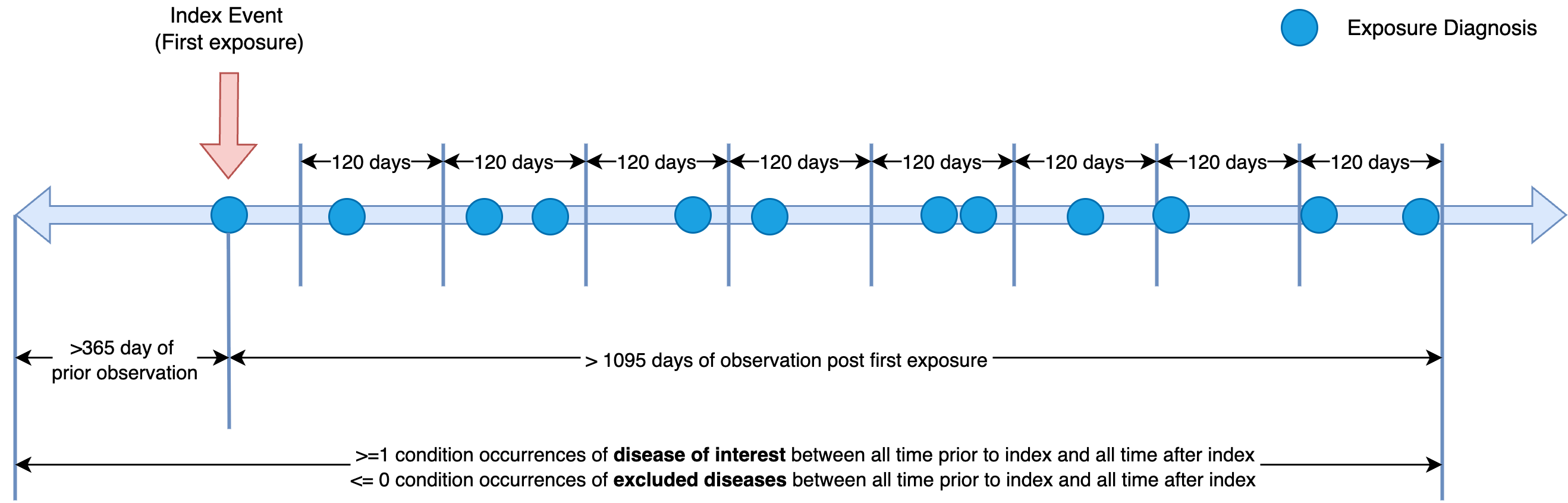}
    \captionsetup{width=0.9\linewidth}
    \caption{Treatment pathway study cohort criteria: Patients were required to have at least 365 days of observation prior to their first drug exposure for the disease of interest. Following this initial exposure, patients had to maintain continuous treatment, defined as having at least one drug exposure every 120 days over a 3-year period.}
    \label{fig:htn_treatment_pathway}
\end{figure*}

To assess the fidelity of the synthetic dataset in capturing complex longitudinal patterns, we replicated the treatment pathway analyses for Hypertension (HTN), Diabetes, and Depression originally proposed by Hripcsak \textit{et al.} \cite{Hripcsak2016}. These analyses impose strict inclusion criteria, such as requiring a three-year follow-up period for HTN patients with continuous exposure to antihypertensive medications across nine consecutive 120-day intervals. As a result, the final cohorts are typically small and highly selective, making them a sensitive test of temporal coherence. The full study design is illustrated in Figure~\ref{fig:htn_treatment_pathway}.

Finally, we evaluated the privacy risks associated with synthetic data sharing using the framework introduced by Yan \textit{et al.} \cite{Yan_Brad_2022}. This framework includes four types of attacks: Membership Inference \cite{Zhang2022}, Attribute Inference \cite{Choi2017}, Meaningful Identity Disclosure \cite{el2020evaluating}, and Nearest Neighbor Adversary Attacks (NNAA) \cite{Wang2018}. We adopted the evaluation metrics and decision thresholds directly from Yan \textit{et al.}, and provide additional implementation details in Supplementary Section~\ref{sec:privacy_evaluation_metrics}.

\subsubsection{Generalization on EHRShot Benchmark}
To evaluate the generalizability of \cehrgpt{}, we utilized the \ehrshot{} benchmarking dataset, which includes 6,739 comprehensive de-identified EHR records from Stanford Medicine \cite{Wornow2023}. This dataset is structured into 15 manually created tasks—both binary and multi-class classification tasks—across three categories: Operational Outcomes (e.g., 30-day readmission), Assignment of New Diagnoses (e.g., 1-year risk of pancreatic cancer), and Anticipating Lab Test Results. Given that \cehrgpt{} was specifically pre-trained on visits, conditions, procedures, and medications, we excluded the Anticipating Lab Test Results category from our analysis.

We opted for the OMOP version of \ehrshot{} for our experiments as both \cehrgpt{} and \gptvanilla{} were trained using the CUIMC OMOP, where medical events are coded using the standard OMOP vocabularies. For each task, we extracted all relevant events prior to the prediction time from the \ehrshot{} OMOP and transformed them into the \cehrgpt{} patient representation. We evaluated both linear probing and full fine-tuning on the downstream tasks, similar to the procedure described in Section~\ref{sec:feature_representation}. In both settings, we extracted the vector corresponding to the final token of the patient sequence from the \cehrgpt{} output and used it to train a linear classifier. The key distinction between the two approaches is that in linear probing, the pre-trained model weights were frozen, whereas in full fine-tuning, all weights were updated during training.  For linear probing, we used the original tokenizer vocabulary and excluded any events not present in it. For fine-tuning, we expanded the tokenizer vocabulary to include over 5000 standard OMOP concept IDs specific to the Stanford EHR data, which were absent in the Columbia EHR. 
We conducted a hyperparameter search over 10 trials, sampling learning rates uniformly between $1\times 10^{-5}$ and $1\times 10^{-4}$. Sample packing was used during training, with a maximum of 16,384 tokens allowed per batch. 

For model comparisons, we included all models listed on the \ehrshot{} leaderboard. Furthermore, we considered the best-performing model from a recent study by the authors of Context Clues, which evaluated several causal language model architectures (\gpt{}, \llama{}, \mamba{}, etc.) with various context windows on Stanford EHR data \cite{Wornow2024}. Their findings, reported on the \ehrshot{}, served as an additional baseline in our experiments.

\section{Experiments and Results}
Our model, \cehrgpt{}, demonstrates versatility across three key capabilities: feature representation, zero-shot prediction, and synthetic data generation. For comparison, we include \motor{} as a SOTA EHR foundation model focused on representation learning. Additionally, we evaluate a variant of \cehrgpt{}, referred to as \gptvanilla{}, which excludes TD and TTE learning objectives. 

\gptvanilla{} is assessed under the same experimental conditions as \cehrgpt{} for the three aforementioned capabilities, except the large-scale synthetic data generation experiment, which was excluded due to the computational cost of generating 3.7 million synthetic patients. Instead, we evaluate its long-term generation ability through a targeted analysis of the treatment pathway using a smaller synthetic cohort of 100,000 patients.

Finally, we assess the generalizability of \cehrgpt{} using the \ehrshot{} benchmarking dataset \cite{Wornow2023}, which provides a diverse range of clinical prediction tasks.


\subsection{Feature Representation}\label{sec:feature_representation}
To evaluate \cehrgpt{}'s ability to generate meaningful patient representations, we assessed its performance under both linear probing and fine-tuning settings, and compared it with \motor{} and \gptvanilla{}. The Receiver Operating Characteristic (ROC) curves are shown in Figure~\ref{fig:roc_curves}, while the Precision-Recall (PR) curves are provided in Supplementary Figure~\ref{fig:pr_curves}. We reported ROC-AUC and PR-AUC across all tasks in Table~\ref{tab:prediction_model_results}.

Under the linear probing setting, \motor{} achieved the highest AUROC and AUPRC across all tasks, outperforming both \cehrgpt{}\textbf{-L} and \gptvanilla{}\textbf{-L}. This may reflect differences in pretraining strategies, as \motor{} is explicitly optimized for representation learning (e.g., time-to-event prediction), whereas \cehrgpt{} is trained using next-token prediction. 

With full model fine-tuning, both \cehrgpt{}\textbf{-T} and \gptvanilla{}\textbf{-T} showed substantial improvements across all tasks compared to their linear probing counterparts. Both models also outperformed \motor{} on several tasks, including \textit{HF Readmission}, \textit{Hospitalization}, \textit{AFib Ischemic Stroke}, and \textit{CAD CABG}. The largest gains were observed on the \textit{AFib Ischemic Stroke} task, where fine-tuning improved AUROC by 7\% and AUPRC by 45\%. 
While \motor{} demonstrated minimal performance gains from fine-tuning in prior work \cite{Steinberg2023}, \cehrgpt{} showed competitive results, particularly when fine-tuned. 

Comparing \gptvanilla{} and \cehrgpt{}, we observed that the incorporation of time token–specific learning objectives had minimal impact on supervised prediction tasks (Table~\ref{tab:prediction_model_results}). However, as described in the next sections, these objectives showed its clear benefits for zero-shot generalization and synthetic data generation.

\begin{figure*}[!htbp]
    \vspace{-2em} 
    \centering
\includegraphics[width=0.97\linewidth]{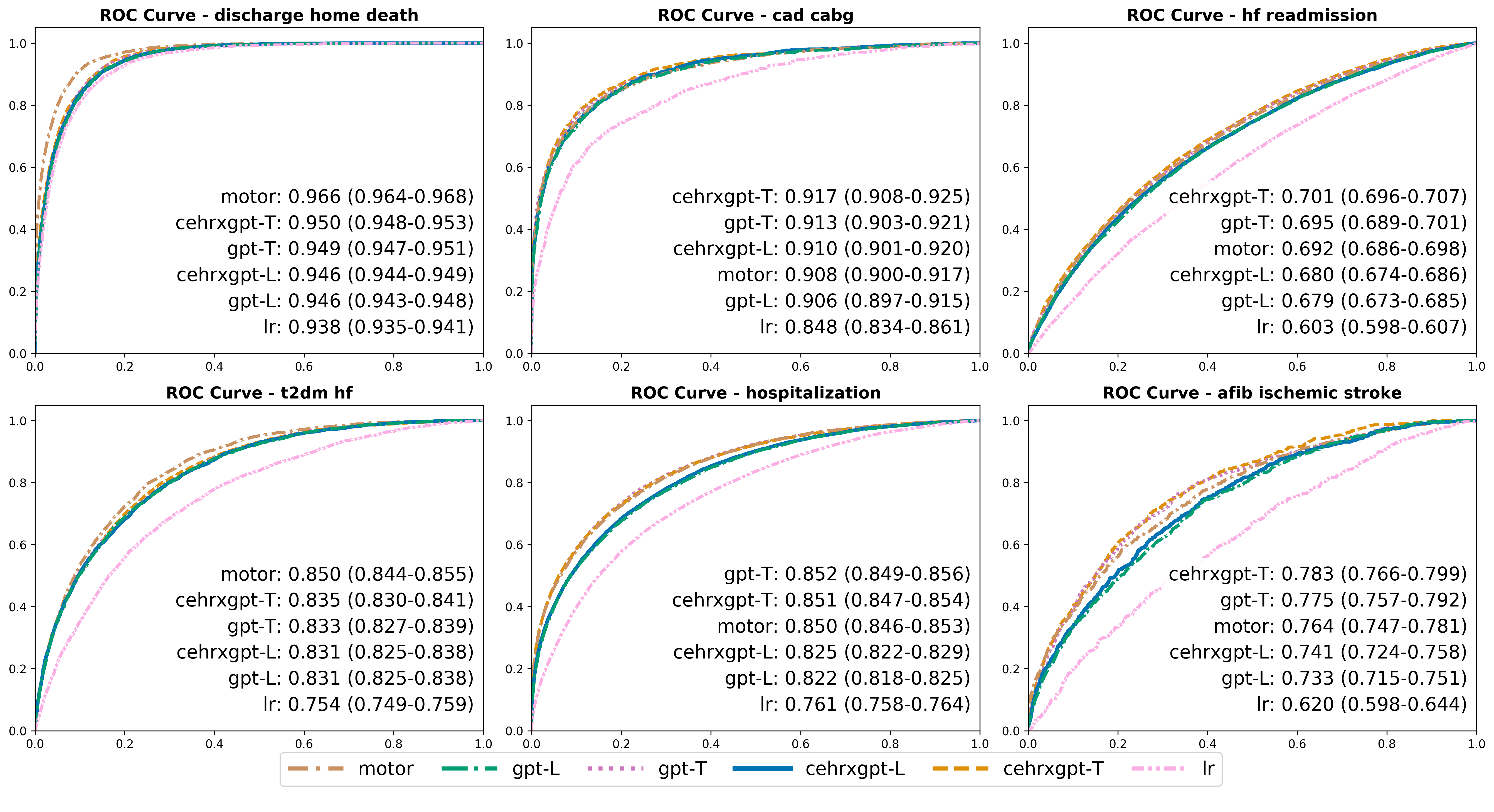}
    \caption{Receiver Operating Characteristic (ROC) curves comparing the performance across six clinical prediction tasks. Models are distinguished by unique color and line style combinations. Performance metrics show Area Under the Curve (AUC) values with 95\% confidence intervals calculated using bootstrap sampling. Models are ranked in descending order of AUC performance within each subplot, with annotations positioned in the bottom right corner. The shared legend displays all evaluated models with their corresponding visual representations. Abbreviations: \cehrgpt{}\textbf{-L} (\cehrgpt{} with linear probing), \cehrgpt{}\textbf{-T} (\cehrgpt{} with fine-tuning), \gptvanilla{}\textbf{-L} (The \cehrgpt{} original variant with linear probing), \gptvanilla{}-\textbf{T} (The \cehrgpt{} original variant with fine-tuning), LR Baseline (Logistic Regression).}
    \label{fig:roc_curves}
    \vspace{-1em} 
\end{figure*}

\begin{table*}[hbt]
\renewcommand{\arraystretch}{1.5}
\begin{tabular}{c|c|ccccccc}
\toprule
\textbf{Cohort} & \textbf{Metric} & 
\makecell{\textbf{LR}\\\textbf{Baseline}} & 
\makecell{\textbf{CEHR}\\\textbf{GPT}-\textbf{L}} & \makecell{\textbf{GPT}-\textbf{L}} & \makecell{\textbf{CEHR}\\\textbf{GPT}-\textbf{T}} & \makecell{\textbf{GPT}-\textbf{T}} &  \textbf{MOTOR}\\
\midrule
\makecell{HF readmission} 
& \makecell{AUROC\\AUPRC}
& \makecell{64.1 $\pm$ 0.3\% \\ 35.1 $\pm$ 0.2\%}
& \makecell{68.0 $\pm$ 0.3\% \\ 37.4 $\pm$ 0.5\%}
& \makecell{67.9 $\pm$ 0.3\% \\ 37.6 $\pm$ 0.5\%}
& \makecell{70.1 $\pm$ 0.3\% \\ 40.5 $\pm$ 0.5\%}
& \makecell{69.5 $\pm$ 0.3\% \\ 39.8 $\pm$ 0.5\%}
& \makecell{69.2 $\pm$ 0.3\% \\ 39.2 $\pm$ 0.5\%} 
\\

\makecell{T2DM HF} 
& \makecell{AUROC\\AUPRC}
& \makecell{77.9 $\pm$ 0.7\% \\ 28.3 $\pm$ 1.3\%}
& \makecell{83.1 $\pm$ 0.3\% \\ 38.8 $\pm$ 0.8\%}
& \makecell{83.1 $\pm$ 0.3\% \\ 38.3 $\pm$ 0.8\%}
& \makecell{83.5 $\pm$ 0.3\% \\ 39.2 $\pm$ 0.8\%}
& \makecell{83.3 $\pm$ 0.3\% \\ 38.2 $\pm$ 0.8\%}
& \makecell{85.0 $\pm$ 0.3\% \\ 40.2 $\pm$ 0.8\%} 
\\

\makecell{Hospitalization} 
& \makecell{AUROC\\AUPRC}
& \makecell{76.2 $\pm$ 0.2\% \\ 20.3 $\pm$ 0.3\%} 
& \makecell{82.5 $\pm$ 0.2\% \\ 31.7 $\pm$ 0.4\%}
& \makecell{82.2 $\pm$ 0.2\% \\ 30.1 $\pm$ 0.4\%}
& \makecell{85.1 $\pm$ 0.2\% \\ 39.2 $\pm$ 0.4\%}
& \makecell{85.2 $\pm$ 0.2\% \\ 38.9 $\pm$ 0.5\%}
& \makecell{85.0 $\pm$ 0.2\% \\ 38.4 $\pm$ 0.5\%}
\\

\makecell{Discharge mortality} 
& \makecell{AUROC\\AUPRC}
& \makecell{93.7 $\pm$ 0.3\% \\ 51.0 $\pm$ 1.7\%}
& \makecell{94.6 $\pm$ 0.1\% \\ 53.6 $\pm$ 0.8\%}
& \makecell{94.6 $\pm$ 0.1\% \\ 53.4 $\pm$ 0.8\%}
& \makecell{95.0 $\pm$ 0.1\% \\ 56.6 $\pm$ 0.8\%}
& \makecell{94.9 $\pm$ 0.1\% \\ 55.5 $\pm$ 0.8\%}
& \makecell{96.6 $\pm$ 0.1\% \\ 70.7 $\pm$ 0.7\%}
\\

\makecell{AFib ischemic stroke} 
& \makecell{AUROC\\AUPRC}
& \makecell{70.5 $\pm$ 1.1\% \\ 9.0 $\pm$ 0.7\%}
& \makecell{74.1 $\pm$ 0.9\% \\ 11.1 $\pm$ 0.9\%}
& \makecell{73.3 $\pm$ 0.9\% \\ 10.6 $\pm$ 0.9\%}
& \makecell{78.3 $\pm$ 0.8\% \\ 14.9 $\pm$ 1.3\%}
& \makecell{77.5 $\pm$ 0.9\% \\ 13.7 $\pm$ 1.1\%}
& \makecell{76.3 $\pm$ 0.9\% \\ 17.3 $\pm$ 1.4\%}
\\

\makecell{CAD CABG} 
& \makecell{AUROC\\AUPRC}
& \makecell{84.8 $\pm$ 0.7\% \\ 33.3 $\pm$ 1.6\%}
& \makecell{91.0 $\pm$ 0.4\% \\ 53.7 $\pm$ 1.4\%}
& \makecell{90.6 $\pm$ 0.5\% \\ 52.5 $\pm$ 1.4\%}
& \makecell{91.7 $\pm$ 0.4\% \\ 56.9 $\pm$ 1.4\%}
& \makecell{91.3 $\pm$ 0.4\% \\ 57.1 $\pm$ 1.4\%}
& \makecell{90.8 $\pm$ 0.5\% \\ 54.6 $\pm$ 1.3\%}
\\
\bottomrule
\end{tabular}
\centering
\setlength{\tabcolsep}{5pt}
\captionsetup{width=.90\textwidth}
\caption{
    Model performance on each disease prediction task. Abbreviations: \cehrgpt{}-L (\cehrgpt{} with linear probing), \cehrgpt{}-T (\cehrgpt{} with fine-tuning), \gptvanilla{}-\textbf{L} (The \cehrgpt{} original variant with linear probing), \gptvanilla{}-\textbf{T} (The \cehrgpt{} original variant with fine-tuning), LR Baseline (Logistic Regression).
}
\label{tab:prediction_model_results}
\end{table*}

\subsection{Zero-Shot Prediction}
We evaluated the zero-shot prediction capability of \cehrgpt{} using three cohorts described in Table \ref{tab:cohort}, representing short-term, mid-term, and long-term risk prediction tasks: \textit{HF Readmission} (30-day risk), \textit{CAD CABG} (1-year risk), and \textit{T2DM HF} (lifetime risk). 

As shown in Table \ref{tab:zero_shot_prediction}, \cehrgpt{} achieved strong zero-shot performance across all three tasks. On the short-term task (\textit{HF Readmission}), it performed comparably to \gptvanilla{}, while demonstrating slightly improved performance on the mid-term and long-term tasks (\textit{CAD CABG} and \textit{T2DM HF}, respectively). These results highlight the benefits of \cehrgpt{}'s temporal modeling capabilities enabled by time tokens and its autoregressive design, which are especially important in longer-horizon prediction. Importantly, the zero-shot performance of \cehrgpt{} (without task-specific fine-tuning) not only rivals the linear probing performance of pretrained models, but also outperforms count-based baselines on both \textit{HF Readmission} and \textit{CAD CABG}, and matches the best baseline on \textit{T2DM HF}. This underscores \cehrgpt{}’s ability to generalize to diverse clinical prediction tasks.

\begin{center}
\renewcommand{\arraystretch}{1.5}
\begin{tabular}{c|cc}

\toprule
\thead{Cohort} & \thead{\cehrgpt{}} & \thead{\gptvanilla{}} \\
\midrule

\makecell{HF readmission \\ (30 day risk)} 
& \makecell{R = 66.8 $\pm$ 0.8\% \\P = 35.1 $\pm$ 1.4\%}
& \makecell{R = 66.8 $\pm$ 0.9\% \\P = 35.8 $\pm$ 1.4\%}
\\

\makecell{T2DM HF \\ (lifetime risk)} 
& \makecell{R = 78.9  $\pm$ 0.9\% \\ P = 27.3  $\pm$ 1.6\%}
& \makecell{R = 78.7  $\pm$ 0.9\% \\ P = 26.9  $\pm$ 1.5\%}
\\

\makecell{CAD CABG \\ (1-year risk)} 
& \makecell{R = 90.9 $\pm$ 1.1\% \\ P = 55.6 $\pm$ 3.3\%}
& \makecell{R = 90.0 $\pm$ 1.2\% \\ P = 54.9 $\pm$ 3.4\%}
\\
\bottomrule
\end{tabular}
\captionof{table}{Zero-shot prediction performance on the short-term prediction task \textit{HF Readmission}, mid-term prediction task \textit{CAD CABG}, long-term prediction task \textit{T2DM HF}. Abbreviations: R (AUROC), P (AUPRC)}
\label{tab:zero_shot_prediction}
\end{center}

\subsection{Synthetic Data Generation} \label{synthetic_data_generation}
Table~\ref{tab:synthetic_prediction_model_results} presents model performance results using synthetic data. For two tasks—\textit{Hospitalization} and \textit{AFib Ischemic Stroke}—models trained entirely on synthetic data performed comparably to those trained on real data when evaluated on the real test set. In contrast, performance for \textit{HF Readmission} and \textit{CAD CABG} was slightly lower with synthetic-only training. However, augmenting the real training set with synthetic data led to modest yet consistent improvements in test performance across all four tasks, relative to models trained on real data alone, suggesting that \cehrgpt{}–generated data can serve as a useful supplement when real data is limited.

\begin{center}
\renewcommand{\arraystretch}{1.5} 
\begin{tabular}{c|c|ccc}
\toprule
\thead{Cohort} & \thead{M} & \thead{Real} & \thead{Syn} & \thead{Syn + Real}\\
\midrule

\makecell{HF \\ Readmit} 
& \makecell{R\\P}
& \makecell{64.1 $\pm$ 0.3\%\\35.1 $\pm$ 0.2\%}
& \makecell{61.6$\pm$ 0.3\% \\32.0 $\pm$ 0.4\%} 
& \makecell{64.5 $\pm$ 0.3\% \\34.5 $\pm$ 0.5\%} \\

\makecell{Hospital-\\ization} 
& \makecell{R\\P}
& \makecell{76.2 $\pm$ 0.2\% \\20.3 $\pm$ 0.3\% } 
& \makecell{76.2 $\pm$ 0.2\% \\22.3 $\pm$ 0.4\% }
& \makecell{77.7 $\pm$ 0.2\% \\23.8 $\pm$ 0.4 \% }  \\

\makecell{AFib IS} 
& \makecell{R\\P}
& \makecell{70.5 $\pm$ 1.1\%\\9.0 $\pm$ 0.7\% } 
& \makecell{70.7 $\pm$ 1.1\%\\9.7 $\pm$ 0.9\% }
& \makecell{71.2 $\pm$ 1.1\%\\10.4 $\pm$ 0.9\% }\\

\makecell{CAD \\ CABG} 
& \makecell{R\\P}
& \makecell{84.8 $\pm$ 0.7\%\\33.3 $\pm$ 1.6\%} 
& \makecell{82.9 $\pm$ 0.7\%\\26.5 $\pm$ 1.5\%}
& \makecell{85.0 $\pm$ 0.7\%\\32.6 $\pm$ 1.6\%} \\
\bottomrule
\end{tabular}
\captionof{table}{Logistic regression model performance (AUROC and AUPRC) across different datasets using synthetic datasets generated from \cehrgpt{}. The columns compare the model performance when 1) trained on synthetic data alone and tested on the real test set, and 2) trained on synthetic + real data, tested on the real test set. Abbreviations: HF Readmit(HF Readmission), AFib IS (AFib Ischemic Stroke), M(Metric), R (AUROC), P (AUPRC)}
\label{tab:synthetic_prediction_model_results}
\end{center}


In treatment pathway analyses, the synthetic dataset generated by \cehrgpt{} closely matched real-world prevalence for HTN and Depression, though it underestimated Diabetes (Supplementary Table~\ref{tab:treatment_pathway_results}): HTN 0.41\% vs. 0.45\%, Diabetes 0.11\% vs. 0.18\%, and Depression 0.18\% vs. 0.14\%. In contrast, the \gptvanilla{} dataset underestimated HTN (0.26\%) and Diabetes (0.10\%) while only aligning on Depression (0.14\%). These findings underscore the value of timeline-aware modeling, with \cehrgpt{} more faithfully preserving long-term patient trajectories than \gptvanilla{}.


Privacy evaluation results are shown in Supplementary Table~\ref{tab:privacy_metrics_results}. Across all four attack types, \cehrgpt{}'s synthetic dataset achieved risk scores below 0.333, a threshold recommended by Yan \textit{et al.} to indicate minimal privacy risk. These results demonstrate that \cehrgpt{} can generate synthetic EHR data with low measurable privacy risk while preserving clinical utility for downstream modeling.

\subsection{Generalization on EHR-SHOT Benchmark}
We demonstrated generalizability of \cehrgpt{} by evaluating it on the external \ehrshot{} benchmarking dataset, which enables comparison with several competitive models available on the \ehrshot{} Leaderboard, including \motor, Context, CLMBR, GBM, LR, RF, and \gptvanilla{} (Table~\ref{tab:metrics} and Supplementary table~\ref{tab:ehrshot_baseline_metrics}). Overall, \cehrgpt{} outperformed \gptvanilla{} in both fine-tuned and linear-probing settings. For \textit{Patient Outcomes}, fine-tuning \cehrgpt{} improved performance, with average gains of 1.2\% in AUROC and 5.0\% in AUPRC compared to linear probing. In contrast, on \textit{New Diagnosis} tasks, fine-tuning \cehrgpt{} yielded a higher AUROC with an average gain of 1.9\% but slightly reduced AUPRC by 0.6\%. Interestingly, \gptvanilla{} followed a different pattern: for \textit{Patient Outcomes}, fine-tuning and linear probing performed similarly, with fine-tuning achieving a modest average improvement of 0.9\% in AUPRC. However, for \textit{New Diagnosis}, fine-tuning degraded performance relative to linear probing, particularly for two low-prevalence cohorts, \textit{Celiac} and \textit{Pancreatic Cancer}. A similar trend was observed for \cehrgpt{} on \textit{Pancreatic Cancer}, where linear probing outperformed fine-tuning. These results suggest that under extreme class imbalance, linear probing may offer a more robust approach for predicting rare events than fine-tuning. 

More importantly, both fine-tuned and linear-probing versions of \cehrgpt{} demonstrated strong performance across these tasks, achieving the best results on \textit{New Diagnosis} prediction compared to all other models. Notably, \cehrgpt{} outperformed models such as \clmber{}, despite the latter being trained on data drawn from the same distribution (Stanford EHR data) as the \ehrshot{} dataset. This suggests that simply extending the context window or adopting alternative architectures does not necessarily yield better performance on these tasks, and that \cehrgpt{}’s novel patient representation design and time-specific learning objectives may be key contributors to its improved performance.

\begin{table*}[htb]
\renewcommand{\arraystretch}{1.3}
\centering
\begin{tabular}{c|c|c|ccccccc}
\toprule
\textbf{Task} & \textbf{Prev} & \textbf{Metric} & \textbf{MOTOR} & \textbf{Context} & \textbf{\makecell{Best\\Baseline}} & \textbf{\makecell{CEHR\\GPT-T}} & \textbf{\makecell{CEHR\\GPT-L}} & \textbf{GPT-T} & \textbf{GPT-L} \\ 
\midrule
\makecell[c]{Patient \\ Outcome}  & 
& 
\makecell{R \\ P} & \makecell{83.6\% \\ 49.0\%} & 
\makecell{83.3\% \\ N/A} & 
\makecell{82.4\% \\ 43.7\%} & 
\makecell{83.3\% \\ 46.7\%} & 
\makecell{82.1\% \\ 41.7\%} & 
\makecell{82.6\% \\ 44.0\%} & 
\makecell{81.7\% \\ 40.8\%}\\

\hline

\makecell[c]{ICU \\ Admission} & 4.5\% & 
\makecell{R \\ P} & \makecell{ \\ } & 
\makecell{ \\ } & 
\makecell{84.8\% \\ 32.4\%} & 
\makecell{85.0$\pm$1.8\% \\ 32.7$\pm$4.9\%} & 
\makecell{84.0$\pm$1.9\% \\ 26.2$\pm$4.4\%} & 
\makecell{83.1$\pm$2.1\% \\ 24.4$\pm$4.8\%} & 
\makecell{84.0$\pm$2.3\% \\ 23.7$\pm$4.7\%}\\

\makecell[c]{Long LOS} & 25.2\% & 
\makecell{R \\ P} & \makecell{ \\ } & \makecell{ \\ } & 
\makecell{81.4\% \\ 57.6\%} & 
\makecell{84.4$\pm$0.1\% \\ 65.3$\pm$2.4\%} & 
\makecell{82.1$\pm$0.1\% \\ 59.9$\pm$2.2\%} & 
\makecell{84.0$\pm$0.1\% \\ 65.1$\pm$2.2\%} & 
\makecell{81.8$\pm$0.1\% \\ 60.0$\pm$2.2\%}\\

\makecell[c]{30-day \\ Readmission} & 13.0\% & 
\makecell{R \\ P} & \makecell{ \\ } & \makecell{ \\ } & 
\makecell{81.0\% \\ 41.2\%} & 
\makecell{80.5$\pm$1.5\% \\ 42.2$\pm$3.3\%} & 
\makecell{80.2$\pm$1.4\% \\ 39.1$\pm$3.1\%} & 
\makecell{80.6$\pm$1.4\% \\ 42.4$\pm$3.1\%} & 
\makecell{79.3$\pm$1.5\% \\ 37.6$\pm$3.4\%} \\

\hline
\makecell[c]{New \\ Diagnosis} &  & \makecell{R \\ P} & 
\makecell{75.6\% \\ 19.3\%} & 
\makecell{75.6\% \\ N/A} & 
\makecell{74.9\% \\ 21.2\%} &  
\makecell{75.9\% \\ 20.8\%} & 
\makecell{74.0\% \\ 21.4\%} & 
\makecell{72.0\% \\ 20.4\%} & 
\makecell{72.8\% \\ 21.5\%} \\
\hline

\makecell[c]{Acute MI} & 6.8\% & \makecell{R \\ P} & 
\makecell{ \\ } & \makecell{ \\ } & 
\makecell{74.7\% \\ 18.4\%} & 
\makecell{76.2$\pm$2.1\% \\ 18.6$\pm$3.3\%} & 
\makecell{73.4$\pm$1.8\% \\ 18.4$\pm$2.9\%} & 
\makecell{75.7$\pm$1.9\% \\ 18.3$\pm$1.9\%} & 
\makecell{76.3$\pm$2.0\% \\ 19.2$\pm$2.6\%} \\

\makecell[c]{Celiac} & 1.3\% & 
\makecell{R \\ P} & \makecell{ \\ } & \makecell{ \\ } & 
\makecell{75.8\% \\ 23.1\%} & 
\makecell{66.8$\pm$5.8\% \\ 1.5$\pm$0.9\%} & 
\makecell{63.6$\pm$5.5\% \\ 3.0$\pm$0.6\%} & 
\makecell{53.9$\pm$3.3\% \\ 1.3$\pm$0.5\%} & 
\makecell{62.4$\pm$8\% \\ 1.1$\pm$1.6\%} \\

\makecell[c]{Pancreatic \\ Cancer} & 3.8\% & 
\makecell{R \\ P} & \makecell{ \\ } & \makecell{ \\ } & 
\makecell{88.5\% \\ 47.2\%} & 
\makecell{83.4$\pm$2.8\% \\ 37.2$\pm$6.2\%} & 
\makecell{83.9$\pm$2.9\% \\ 38.7$\pm$6.2\%} & 
\makecell{77.7$\pm$3.9\% \\ 35.9$\pm$6.9\%} & 
\makecell{81.6$\pm$3.3\% \\ 41.5$\pm$6.8\%} \\

\makecell[c]{Hypertension} & 13.7\% & 
\makecell{R \\ P} & \makecell{ \\ } & \makecell{ \\ } & 
\makecell{71.8\% \\ 25.8\%} & 
\makecell{73.6$\pm$2.0\% \\ 28.5$\pm$3.0\%} & 
\makecell{71.4$\pm$2.2\% \\ 26.2$\pm$3.0\%} & 
\makecell{72.9$\pm$2.3\% \\ 28.8$\pm$3.4\%} & 
\makecell{70.6$\pm$2.2\% \\ 24.7$\pm$3.0\%} \\

\makecell[c]{Lupus} & 2.2\% & 
\makecell{R \\ P} & \makecell{ \\ } & \makecell{ \\ } & 
\makecell{79.3\% \\ 17.9\%} & 
\makecell{80.6$\pm$4.2\% \\ 8.5$\pm$7.7\%} & 
\makecell{79.1$\pm$5.2\% \\ 10.1$\pm$3.9\%} & 
\makecell{76.1$\pm$6.6\% \\ 6.0$\pm$0.5\%} & 
\makecell{74.5$\pm$3.5\% \\ 6.$\pm$3.6\%} \\

\makecell[c]{Hyperlipidemia} & 12.7\%  & 
\makecell{R \\ P} & \makecell{ \\ } & \makecell{ \\ } & 
\makecell{72.0\% \\ 25.8\%} & 
\makecell{74.7$\pm$2.1\% \\ 30.4$\pm$3.4\%} & 
\makecell{72.6$\pm$1.9\% \\ 31.7$\pm$3.2\%} & 
\makecell{75.7$\pm$1.9\% \\ 32.3$\pm$3.2\%} & 
\makecell{71.4$\pm$1.9\% \\ 27.3$\pm$3.4\%} \\
\hline
\end{tabular}
\caption{Performance metrics of selected models for patient outcome and new diagnosis predictions using the \ehrshot{} dataset. This table summarizes both individual and average model performance across tasks. The \emph{Best Baseline} column shows the best metric among CLMBR, GBM, LR, and RF. For \motor{}, only aggregated performance data is available since specific task details were not published on the \ehrshot{} leaderboard. For entries marked as 'Context,' only the highest AUROC is reported due to the absence of AUPRC metrics in the original publication. Abbreviations: \cehrgpt{}-L (\cehrgpt{} with linear probing), \cehrgpt{}-T (\cehrgpt{} with fine-tuning), \gptvanilla{}-\textbf{L} (The \cehrgpt{} original variant with linear probing), \gptvanilla{}-\textbf{T} (The \cehrgpt{} original variant with fine-tuning). Prev (Prevalence), R (AUROC), P (AUPRC)}
\label{tab:metrics}
\end{table*}

\section{Discussion}
\cehrgpt{} is, to our knowledge, the first EHR foundation model comprehensively evaluated across three core capabilities: feature representation, zero-shot prediction, and synthetic data generation. While prior models typically specialize in one area, \cehrgpt{} demonstrates broad versatility, enabling a wide range of clinical tasks without requiring task-specific retraining. This flexibility is particularly valuable in real-world settings with limited labeled data, evolving populations, or rapidly changing outcome definitions.  

A key enabler of this versatility is \cehrgpt{}’s use of time tokens, along with Time Decomposition (TD) and Time-to-Event (TTE) objectives. These mechanisms help encode temporal structure directly into patient sequences and appear to enhance generalizability—particularly in zero-shot and transfer settings. For example, in the external \ehrshot{} benchmark, \cehrgpt{} achieved strong performance through vocabulary expansion and full fine-tuning. It matched top-reported models on \textit{Patient Outcome} tasks and outperformed all previously reported models on \textit{New Diagnosis} tasks. While \motor{}, \clmber{}, \llama{}, and \mamba{} were evaluated using linear probing in prior studies, their performance could improve with fine-tuning.

In contrast to \motor{}, which consistently outperformed other models in linear probing, \cehrgpt{}’s representation performance was more modest. This may reflect differences in pretraining: \motor{} was trained to forecast thousands of future events across time intervals, promoting long-range structure learning. \cehrgpt{}, by comparison, was trained with next-token prediction and auxiliary temporal objectives—favoring sequential modeling over embedding quality. These results suggest that the alignment between pretraining and downstream tasks plays a critical role in determining the effectiveness of linear probing. Another factor is feature coverage: \cehrgpt{} currently excludes measurements and observations, domains used by \motor{}. Expanding \cehrgpt{}’s input domains may help close this gap.

Interestingly, \gptvanilla{} (\cehrgpt{} without time-specific learning objectives) demonstrated similar performance to \cehrgpt{} on in-distribution representation tasks, but underperformed in zero-shot prediction, treatment pathway analysis using synthetic data, and external validation. This suggests that time tokens act as a form of structural regularization—limiting overfitting to the training distribution while improving temporal generalization. Analogous to regularization in traditional models, the TD and TTE objectives guide the model to learn more transferable temporal representations, which is critical for generalization across health systems.

It is also worth emphasizing that feature representation and zero-shot prediction serve different clinical purposes. Representation learning supports embedding-based tasks like clustering or risk stratification, while zero-shot prediction enables outcome forecasting without retraining. In settings where labeled data or retraining resources are scarce—such as low-resource hospitals, emerging diseases, or real-time triage—zero-shot capabilities may offer faster and more scalable deployment.

In the context of synthetic data generation, most prior frameworks—such as Theodorou \textit{et al.}~\cite{Theodorou2023}—adopt a conditional generation strategy to synthesize patients based on predefined conditions like \textit{T2DM}. However, this approach relies on condition-specific labels assigned in advance and does not generalize well to novel or unseen conditions. In contrast, our method models the full sequence distribution of the Columbia patient population using a specialized representation that preserves complete patient timelines. A key distinction of our approach is its ability to convert generated sequences back into a common data model, enabling researchers to identify patients with any condition post hoc using standard tools such as pandas or SQL. 

Taken together, these results position \cehrgpt{} as a flexible and general-purpose foundation model for EHR data. Its support for zero-shot prediction, synthetic data generation, and representation learning makes it well suited for diverse clinical workflows, including cohort discovery, surveillance, and rapid model prototyping. Future work may extend \cehrgpt{} by incorporating additional domains (e.g., measurements, labs, observations), improving computational efficiency (e.g., via adapters or distillation), and enhancing clinical interpretability through human-in-the-loop evaluation.

\section{Data availability}
Due to institutional policies regarding the sharing of patient-level EHR data, we are unable to share the dataset used for training. However, all code, along with instructions to replicate our work, is accessible on GitHub at \href{https://github.com/knatarajan-lab/cehrgpt}{https://github.com/knatarajan-lab/cehrgpt}. This will enable institutions with an OMOP database to train models and generate synthetic data within their internal environment.

\section{Acknowledgement}
This work was supported by the National Institutes of Health (5U2COD023196, 3OT2OD026556, 5U01TR002062, 5R35GM147004). The funding sources supported individuals in curating and analyzing electronic health record data, which spurred the idea of this work.

\section{Author contributions}
CP conceived the study. CP led the conceptualization and methodology design with contributions from XJ, CP. CP and XJ conducted the data curation and cohort design. CP, XJ carried out the formal analysis, including various evaluation metrics. CP led the project administration. CP, JP, and XJ authored the original draft, which was reviewed and edited by KN, SJ, and NE. All authors had final responsibility for the decision to submit for publication.

\bibliography{main}
\end{multicols}
\newpage
\section{Supplementary Materials}
\subsection{Simulation Study} \label{simulation_study}
We designed a simulation based on the Transformer Encoder \cite{Devlin2019} to address the scenario described in Section~\ref{time_comparison}. To recapitulate, we consider a sequence of two events, where \(x_1, x_2 \in \{0, 1\}\) and \(t_1, t_2\) are their respective timestamps with \(0 \leq t_1 \leq t_2 \leq 28\). We devised the following rules for constructing the output \(y\):

\begin{enumerate}
    \item If \( (t_2 - t_1) \mod 4 = 0\) and \(x_1 = 1\), then \(y = \neg x_2\).
    \item Else if \( (t_2 - t_1) \mod 3 = 0\) and \(x_1 = 0\), then \(y = x_2\).
    \item Else if \(t_2 - t_1 \leq 7\), then \(y = \text{XOR}(x_1, x_2)\).
    \item Else if \(7 < t_2 - t_1 \leq 14 \), then \(y = \text{AND}(x_1, x_2)\).
    \item Else if \(t_2 - t_1 > 14\), then \(y = \text{OR}(x_1, x_2)\).
\end{enumerate}

\begin{figure}[H]
    \centering
    \includegraphics[width=0.6\linewidth]{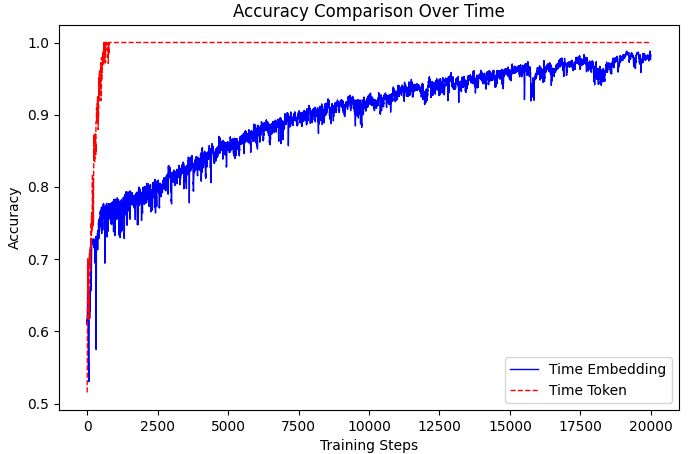}
    \captionsetup{width=0.8\linewidth}
    \caption{The accuracy comparison between the model using the time tokens and the model summing the input embeddings and time embeddings.}
    \label{accuracy_comparison}
\end{figure}

We generated 1,000 simulated data points using uniform sampling and constructed the corresponding output \(y\). For illustration, we used the Transformer encoder architecture to develop two models: one employing the summation strategy (\(Model_{sum}\)) and the other incorporating a time token (\(Model_{timetoken}\)) between \(x_1\) and \(x_2\), utilizing the time interval \(\Delta t = t_2 - t_1\). We configured both models with an embedding size of 16 and 2 encoder layers with a hidden size of 16 and an intermediate size of 32. All dropout rates were disabled to maintain simplicity in the model architecture. The conceptual configurations of these models are outlined as follows:

\begin{minipage}{0.45\textwidth}
\[
\begin{aligned}
    & \textit{Model with summation:} \\
    & e_{x1}, e_{x2}, e_{t1}, e_{t2} = \text{Embedding}(x_1, x_2, t_1, t_2), \\
    & y_{sum} = \text{linear}(\text{Encoder}(e_{x1} + e_{t1}, e_{x2} + e_{t2})) \\\\
\end{aligned}
\]
\end{minipage}
\hfill
\begin{minipage}{0.45\textwidth}
\[
\begin{aligned}
    & \textit{Model with time tokens:} \\
    & \Delta t = t_2 - t_1, \\
    & e_{x1}, e_{x2}, e_{\Delta t} = \text{Embedding}(x_1, x_2, \Delta t), \\
    & y_{timetoken} = \text{linear}(\text{Encoder}(e_{x1}, e_{\Delta t}, e_{x2})) \\
    \\
\end{aligned}
\]
\end{minipage}
We trained each model for 20,000 steps, utilizing a batch size of 128. To monitor convergence, we assessed the model's performance at every 100-step interval on the entire training dataset, which consisted of 1,000 data points. As depicted in Figure~\ref{accuracy_comparison}, the model employing time tokens exhibited rapid convergence, achieving perfect accuracy (1.0). In contrast, the accuracy of the model using the summation strategy improved gradually over time and failed to converge even after 20,000 steps. While \(Model_{sum}\) could potentially resolve the constructed function in the simulation by either extending its training duration or increasing its model size, it is clear that \(Model_{token}\) demonstrates greater flexibility in managing time-dependent tasks. 

In this simulation, the transformer models utilizing the summation strategy struggled to achieve rapid convergence, raising concerns about their capability to handle the more intricate patterns observed in patient trajectories within EHR data. The practice of merely adding time embeddings to concept embeddings does not appear to be the most effective approach for modeling longitudinal EHR records. This assertion is supported by an ablation study from our previous work \cite{Pang2021}, where the performance of the BERT model significantly deteriorated after the removal of time tokens from the patient sequences, whereas the elimination of the time embeddings had a less pronounced impact. 

\subsection{Model Training}
Below, we summarize the pretraining configurations used for each model. For \motor{}, we adopted the optimal hyperparameters reported by the original authors~\cite{Steinberg2023}. For \cehrgpt{} and \gptvanilla{}, we followed the parameter settings described in Pang \textit{et al.}~\cite{Pang2024}. All models were trained for up to 10 epochs with an early stopping criterion: training was halted if the evaluation loss did not improve by more than 0.1\%. \motor{} used 5\% of the training data for validation, while \cehrgpt{} and \gptvanilla{} used 10\%.

\begin{table}[H]
\renewcommand{\arraystretch}{1.5}
\centering
\begin{tabular}{c|ccc}
\hline
 & \thead{\motor{}} & \thead{\gptvanilla{}} & \thead{\cehrgpt{}} \\
\hline
\makecell{No. \\Parameters} & ~120M & ~130M & ~130M \\
\makecell{Learning \\ Rate} & $1\times10^{-5}$ & $1\times 10^{-3}$ & $1\times 10^{-3}$ \\
\makecell{Optimizer} & AdamW & AdamW & AdamW \\
\makecell{Weight Decay} & 0.1 & 0.01 & 0.01 \\
\makecell{Adam Beta1} & 0.9 & 0.9 & 0.9 \\
\makecell{Adam Beta2} & 0.95 & 0.999 & 0.999 \\
\makecell{Warmup steps} & 500 & 500 & 500 \\
\hline
\end{tabular}
\captionsetup{width=0.8\linewidth}
\caption{Summary of the model architectures}
\label{tab:model_summary}
\end{table}

\newpage

\subsection{Feature Representation Results}

\begin{figure*}[!htbp]
    \centering
\includegraphics[width=0.97\linewidth]{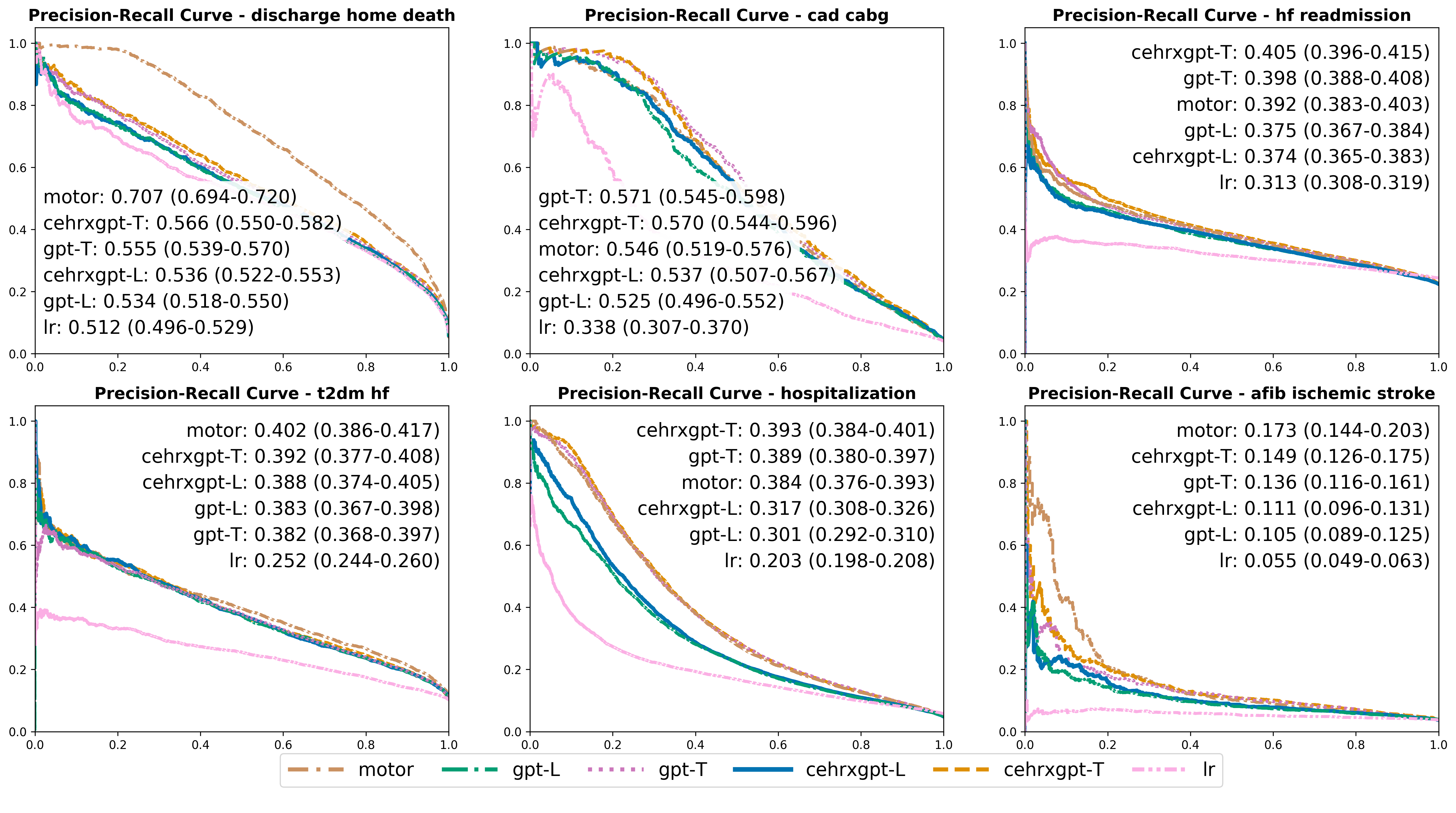}
    \caption{Precision-Recall (PR) curves comparing model performance across six clinical prediction tasks. Models are distinguished by unique color and line style combinations. Performance metrics show Precision-Recall Area Under the Curve (PR AUC) values with 95\% confidence intervals calculated using bootstrap sampling. Models are ranked in descending order of PR AUC performance within each subplot. For most cohorts, annotations are positioned in the top right corner; for \textit{CAD CABG} and \textit{Discharge Home Death} cohorts, annotations are positioned in the bottom left corner to avoid overlap with the curves. The shared legend displays all evaluated models with their corresponding visual representations. Abbreviations: \cehrgpt{}-L (\cehrgpt{} with linear probing), \cehrgpt{}-T (\cehrgpt{} with fine-tuning), \gptvanilla{}-L (The \cehrgpt{} original variant with linear probing), \gptvanilla{}-T (The \cehrgpt{} original variant with fine-tuning), LR Baseline (Logistic Regression).}
    \label{fig:pr_curves}
    \vspace{-1em} 
\end{figure*}

Figure~\ref{fig:pr_curves} presents the Precision-Recall curves for all evaluated models across six clinical tasks, as defined in Table~\ref{Cohort Definition}. Under the linear probing setting, \motor{} consistently outperforms the other models, achieving the highest PR-AUC across all tasks. \gptvanilla{} and \cehrgpt{} closely follow \motor{} on most tasks, with the exception of \textit{Discharge home death}, where the performance gap is more pronounced. In contrast, when fine-tuning is applied, both \gptvanilla{} and \cehrgpt{} show consistent improvements and become the top-performing models in terms of PR-AUC across nearly all tasks, again with the exception of \textit{Discharge home death}.

\newpage
\subsection{Zero Shot Prediction Setup}\label{sec:zero_shot_setup}
Below, we provide two examples illustrating how zero-shot prediction is configured using \cehrgpt{}. Each task requires specifying a list of outcome events using OMOP concept IDs, along with the start and end points of the prediction window. Additionally, the optional parameter \textbf{include\_descendants} can be set to \textbf{True} to leverage the OMOP vocabulary and automatically include all descendant concepts of the specified outcome events. 

\begin{lstlisting}[language=YAML, caption={Zero-shot prediction configuration for 30-day readmission task.}, label={lst:yaml_config}]
task_name: "30_day_readmission_prediction"
outcome_events: ["9201", "262", "8971", "8920"]
include_descendants: false
prediction_window_start: 0
prediction_window_end: 30
max_new_tokens: 128
\end{lstlisting}

\begin{lstlisting}[language=YAML, caption={Zero-shot prediction configuration for CABG outcome prediction task.}, label={lst:cabg_yaml_config}]
task_name: "cabg_prediction"
outcome_events: [
    "43528001",
    "43528003",
    "43528004",
    "43528002",
    "4305852",
    "4168831",
    "2107250",
    "2107216",
    "2107222",
    "2107231",
    "4336464",
    "4231998",
    "4284104",
    "2100873"
]
prediction_window_start: 0
prediction_window_end: 365
max_new_tokens: 1024
include_descendants: true
\end{lstlisting}

\newpage

\subsection{Synthetic Dataset Evaluation}\label{sec:synthetic_data_evaluation}
We generated a total of 3,866,432 synthetic patient sequences, enforcing a minimum sequence length of 20 tokens to ensure the usefulness of the data. Of these, 3,740,307 (96.7\%) were successfully converted back to the OMOP format. Table~\ref{tab:synthetic_statistics} presents a comparison of basic summary statistics between the synthetic and real datasets. To ensure a fair comparison, all real patient sequences shorter than 20 tokens were excluded.

\begin{table}[H]
\renewcommand{\arraystretch}{1.2}
\centering
\begin{tabular}{cc|cc}
\hline
\thead{Category} & \thead{Metric} & \thead{Synthetic Data} & \thead{Source Data} \\
\hline
\multirow{2}{*}{\makecell{Demographics}} 
  & Age (Median) & 53 & 59  \\
  & Female & 54.8\% & 55.7\% \\
\hline
\multirow{3}{*}{\makecell{Visit \\ Length}} 
  & Q1 & 4  & 4  \\
  & Q2 & 8  & 9  \\
  & Q3 & 21 & 27 \\
\hline
\multirow{3}{*}{\makecell{Number of \\ Tokens}} 
  & Q1 & 37  & 37  \\
  & Q2 & 78  & 80  \\
  & Q3 & 212 & 211 \\
\hline
\end{tabular}
\captionsetup{width=0.8\linewidth}
\caption{Comparison of demographic and clinical summary statistics between synthetic and source cohorts.}
\label{tab:synthetic_statistics}
\end{table}

\subsubsection{Concept Prevalence Comparison}
\begin{figure}[H]
    \centering
    \includegraphics[width=0.7\linewidth]{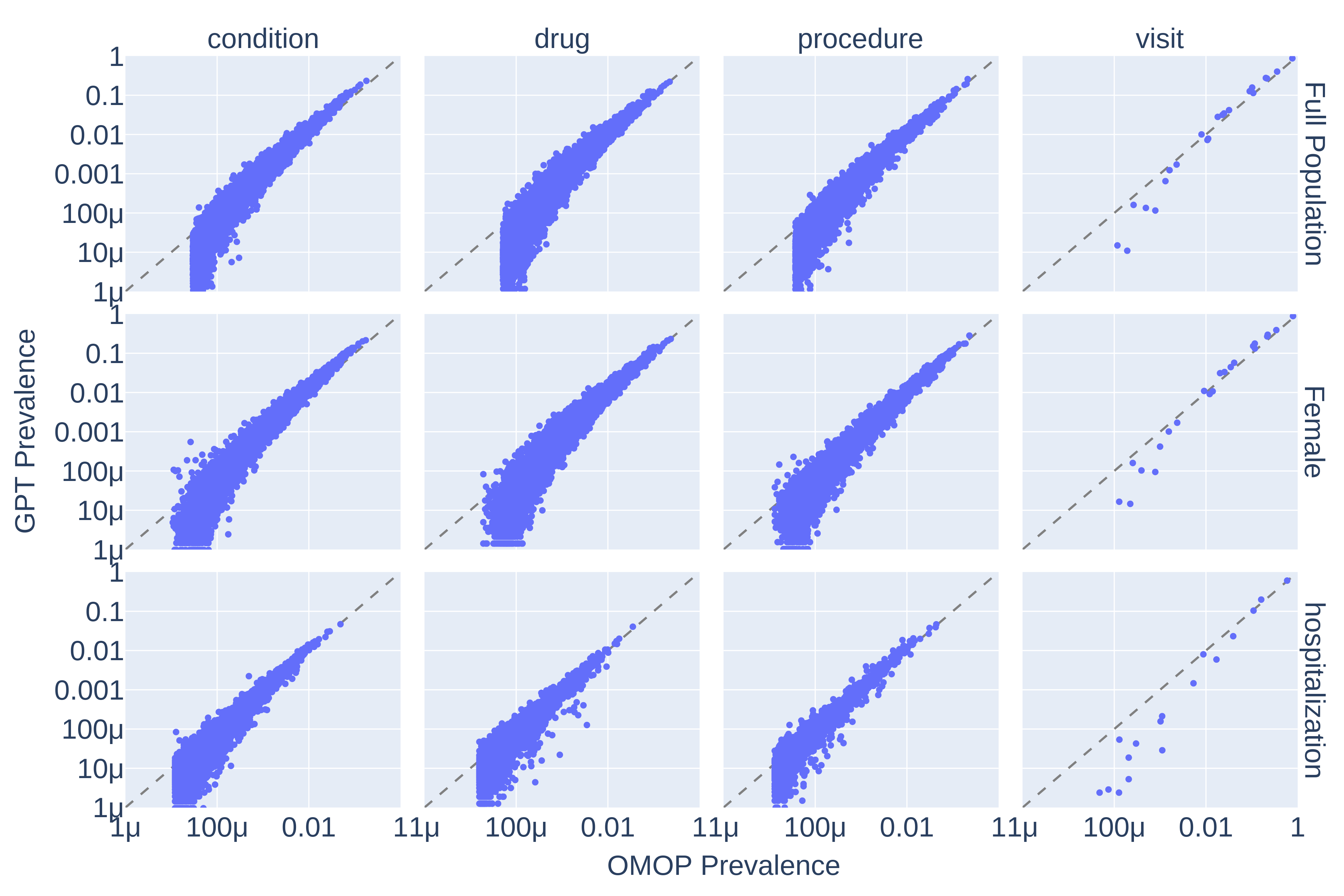}
    \captionsetup{width=0.8\linewidth}
    \caption{The concept prevalence comparison between the real OMOP and the synthetic datasets stratified by domains (condition, drug, procedure, and visit) and populations (full, female, and hospitalization cohorts. Each dot represents an OMOP concept.}
    \label{concept_prevalence_comparison}
\end{figure}
As shown in Figure~\ref{concept_prevalence_comparison}, concept prevalence was reasonably well replicated in the synthetic data, with most points aligning along the diagonal across different domains and populations. High-frequency concepts, in particular, showed strong agreement between the real and synthetic datasets. However, low-frequency concepts tended to be over-represented in the synthetic data, suggesting that the model may amplify rare events.

\subsubsection{Replication of Rare Condition}
We used HIV as a case study to demonstrate that the synthetic data can effectively replicate low-prevalence diseases. Using the OMOP concept ID \href{https://athena.ohdsi.org/search-terms/terms/439727}{439727} (Human immunodeficiency virus infection) and its descendant concepts, we identified 28,601 HIV patients (0.7\%) in the synthetic OMOP dataset and 25,628 HIV patients (0.8\%) in the real dataset. From both cohorts, we extracted the 15 most frequently used HIV treatment drugs and calculated their prevalence within each cohort. As shown in Table~\ref{tab:hiv_drug_prevalence}, the synthetic cohort preserved the ranking and relative ordering of drug prevalence, closely resembling that of the real cohort. However, the prevalence values in the synthetic data were slightly lower overall, which may reflect the inherent difficulty in replicating low-frequency patterns in generative models.

\begin{table}[H]
\centering
\begin{tabular}{c|cc}
\hline
\thead{\makecell{Concept Name}} & \thead{\makecell{Synthetic Prevalence}} & \thead{\makecell{Source Prevalence}} \\
\hline
emtricitabine         & 35.8\% & 39.5\% \\
\makecell{tenofovir alafenamide} & 17.3\% & 23.7\% \\
ritonavir             & 10.8\% & 19.2\% \\
lamivudine            & 10.9\% & 19.2\% \\
dolutegravir          & 11.7\% & 16.7\% \\
abacavir              & 8.1\%  & 14.1\% \\
cobicistat            & 8.9\%  & 13.1\% \\
bictegravir           & 10.8\% & 12.8\% \\
darunavir             & 5.8\%  & 10.8\% \\
efavirenz             & 7.0\%  & 10.4\% \\
elvitegravir          & 7.0\%  & 9.2\%  \\
rilpivirine           & 4.2\%  & 7.4\%  \\
atazanavir            & 3.9\%  & 7.3\%  \\
zidovudine            & 4.0\%  & 6.8\%  \\
lopinavir             & 2.7\%  & 5.2\%  \\
\hline
\end{tabular}
\captionsetup{width=0.8\linewidth}
\caption{HIV-specific drug prevalence comparison between synthetic and source cohorts.}
\label{tab:hiv_drug_prevalence}
\end{table}

\subsubsection{Replication of treatment path studies}
\begin{center}
\renewcommand{\arraystretch}{1.2}
\begin{tabular}{c|c|cc}
\toprule
\thead{Cohort} & \thead{Real} & \thead{\makecell{\cehrgpt{}}} & \thead{\gptvanilla{}} \\
\midrule
Hypertension & 0.45\% & 0.41\% & 0.36\% \\
Diabetes     & 0.18\% & 0.11\% & 0.10\% \\
Depression   & 0.14\% & 0.18\% & 0.14\% \\
\bottomrule
\end{tabular}
\captionof{table}{Prevalence of treatment pathway cohorts, including Hypertension, Diabetes, and Depression, in the real dataset and in synthetic datasets generated using \cehrgpt{} and \gptvanilla{}.}
\label{tab:treatment_pathway_results}
\end{center}

\newpage

\subsubsection{Privacy Evaluation Metric Definitions}\label{sec:privacy_evaluation_metrics}

\begin{table*}[!hbt]
  \centering
  \begin{tabular}{c|c|cccccc}
  \toprule 
  \thead{Privacy Metric}  & \thead{\cehrgpt{}} & \thead{medGAN} & \thead{medBGAN} & \thead{EMR-WGAN} & \thead{WGAN} & \thead{DPGAN} \\
  \midrule
Attribute Inference & 0.027 & 0.0078 & 0.0117 & 0.0680 & 0.0042 & 0.0136 \\
    Membership Inference & 0.1266 & 0.1506 & 0.1828 & 0.2966 & 
    0.1758 &
    0.0000 \\
    Meaningful Identity Disclosure &  0.002 & 0.0021 & 
    0.0027 &
    0.00361 &  0.0034 & 0.0004 \\
    NNAA Risk &  -0.0011 & 0.0008 &  0.0004 & 0.0198 &  0.0085 & 0.0017\\
\hline
  \end{tabular}
  \captionsetup{width=.90\textwidth}
  \caption{Privacy metrics computed on the synthetic data generated by \cehrgpt{}. For reference, we include benchmark values reported by Yan \textit{et al.}~\cite{Yan_Brad_2022}, which were computed on a different dataset (UW). These cited values are not directly comparable but are included as indicative reference points for low privacy risk. For all metrics, lower scores indicate better privacy protection.}
  \label{tab:privacy_metrics_results}
\end{table*}

\noindent
\textbf{Membership Inference Attack}: In a membership inference attack, the goal is to determine whether a real patient record was used in training a generative model by analyzing its synthetic outputs. These attacks are possible when the model overfits to the training data, causing the synthetic data to resemble the training set more closely than data not seen during training. \cite{Zhang2022}
\\
\\
\textbf{Attribute Inference Attack}: An attack in which an adversary, given partial information about a target individual (e.g., demographics and common clinical conditions), attempts to infer sensitive attributes by identifying the most similar patient in the synthetic dataset and using their sensitive information as a proxy \cite{Choi2017}.
\\
\\
\textbf{Meaningful Identity Disclosure Attack}: quantifies the likelihood that a synthetic record can be linked to an individual in an identified population dataset, potentially revealing sensitive information about that person \cite{el2020evaluating}.
\\
\\
\textbf{Nearest Neighbor Adversary Attack}: measures model overfitting by evaluating whether synthetic records are closer to training data than to held-out evaluation data. A smaller distance to training records suggests potential memorization and overfitting by the generative model \cite{Wang2018}.

\newpage

\subsection{Baseline Model Performance on \ehrshot{}}\label{ehrshot_generalization}
Below, we report the performance of the baseline model in the patient outcomes and the new diagnosis tasks from the \ehrshot{} Leaderboard. 

\begin{table*}[htb]
\renewcommand{\arraystretch}{1.3}
\centering
\begin{tabular}{c|c|cccc}
\hline
\thead{Task} & \thead{Metric} & \thead{CLMBR} & \thead{GBM} & \thead{LR} & \thead{RF} \\ \hline
\makecell[c]{Patient \\ Outcome}  & \makecell{AUROC \\ AUPRC} & \makecell{82.4\% \\ 43.7\%} & \makecell{77.4\% \\ 39.4\%} & \makecell{71.9\% \\ 32.2\%} & \makecell{75.1\% \\ 38.1\%} \\
\hline
\makecell[c]{ICU Admission}    & \makecell{AUROC \\ AUPRC} & \makecell{84.8\% \\ 32.4\%} & \makecell{79.9\% \\ 32.4\%} & \makecell{70.1\% \\ 32.4\%} & \makecell{72.1\% \\ 32.4\%} \\

\makecell[c]{Long LOS}         & \makecell{AUROC \\ AUPRC} & \makecell{81.4\% \\ 57.6\%} & \makecell{78.3\% \\ 53.2\%} & \makecell{70.4\% \\ 38.4\%} & \makecell{75.8\% \\ 48.3\%} \\

\makecell[c]{30-day Readmission} & \makecell{AUROC \\ AUPRC} & \makecell{81.0\% \\ 41.2\%} & \makecell{74.1\% \\ 32.7\%} & \makecell{75.1\% \\ 25.8\%} & \makecell{77.5\% \\ 33.5\%} \\

\hline
\makecell[c]{New \\ Diagnosis}    & \makecell{AUROC \\ AUPRC} & \makecell{70.7\% \\ 16.0\%} & \makecell{71.9\% \\ 21.2\%} & \makecell{74.9\% \\ 17.9\%} & \makecell{68.4\% \\ 18.6\%} \\
\hline
\makecell[c]{Acute MI}         & \makecell{AUROC \\ AUPRC} & \makecell{72.9\% \\ 18.3\%} & \makecell{72.5\% \\ 18.4\%} & \makecell{67.8\% \\ 13.5\%} & \makecell{74.1\% \\ 16.5\%} \\

\makecell[c]{Celiac}           & \makecell{AUROC \\ AUPRC} & \makecell{55.7\% \\ 1.7\%} & \makecell{72.3\% \\ 5.8\%} & \makecell{75.8\% \\ 23.1\%} & \makecell{63.9\% \\ 4.5\%} \\

\makecell[c]{Pancreatic Cancer} & \makecell{AUROC \\ AUPRC} & \makecell{81.3\% \\ 25.2\%} & \makecell{82.4\% \\ 37.2\%} & \makecell{85.6\% \\ 15.7\%} & \makecell{88.5\% \\ 47.2\%} \\

\makecell[c]{Hypertension}     & \makecell{AUROC \\ AUPRC} & \makecell{71.8\% \\ 25.8\%} & \makecell{63.7\% \\ 22.1\%} & \makecell{68.9\% \\ 21.2\%} & \makecell{62.7\% \\ 22.6\%} \\

\makecell[c]{Lupus}            & \makecell{AUROC \\ AUPRC} & \makecell{74.7\% \\ 2.4\%} & \makecell{70.3\% \\ 17.9\%} & \makecell{79.3\% \\ 9.3\%} & \makecell{58.7\% \\ 1.6\%} \\

\makecell[c]{Hyperlipidemia}   & \makecell{AUROC \\ AUPRC} & \makecell{67.5\% \\ 22.5\%} & \makecell{69.9\% \\ 25.8\%} & \makecell{72.0\% \\ 24.7\%} & \makecell{62.5\% \\ 19.3\%} \\
\hline
\end{tabular}
\caption{Performance metrics of baseline models for patient outcome and new diagnosis predictions using the \ehrshot{} dataset. This table summarizes both individual and average model performance across tasks. Abbreviations: LR (Logistic Regression), RF (Random Forest), GBM (Gradient Boosting Machine).}
\label{tab:ehrshot_baseline_metrics}
\end{table*}

\end{document}